\documentclass[research]{fcs}
\usepackage{layout}
\usepackage{subfig}
\usepackage{hyperref}
\usepackage{colortbl}
\usepackage{amsthm}
\usepackage{multirow}
\usepackage{amsthm}
\usepackage{pifont}

\usepackage{bbding}
\usepackage{color}
\usepackage{amsmath}

\title{Proxy Robustness in Vision Language Models is Effortlessly Transferable}

\author[1,2]{Xiaowei~Fu}
\author[1,3]{Fuxiang~Huang}
\author[1,2, +]{Lei~Zhang}

\address[1]{Chongqing Key Laboratory of Bio-perception and Multimodal Intelligent
Information Processing, Chongqing University, Chongqing 401331, China}
\address[2]{School of Microelectronics
and Communication Engineering, Chongqing University, Chongqing 401331,
China}
\address[3]{School of Data Science, Lingnan University, Hong Kong}

\corremail{leizhang@cqu.edu.cn}
\fcssetup{
  received = {month dd, yyyy},
  accepted = {month dd, yyyy},
}

\begin{abstract}
As a pivotal technique for improving the defense of deep models, adversarial robustness transfer via distillation has demonstrated remarkable success in conventional image classification tasks. However, this paradigm encounters critical challenges when applied to vision-language models (VLM) (e.g., CLIP): constructing adversarially robust teacher for large-scale multi-modal models demands prohibitively high computational resources. We bridge this gap by revealing an interesting phenomenon: vanilla CLIP (without adversarial training) exhibits intrinsic defensive capabilities against adversarial examples generated by another CLIP with different architectures. 
We formally define this as \textbf{proxy adversarial robustness}, and naturally propose a Heterogeneous Proxy Transfer (HPT) framework that establishes cross-architectural robustness distillation channels between CLIP variants, effortlessly enabling the VLM robustness transfer from proxy to target models. Yet, such proxy transfer paradigm easily induces severe overfitting, leading to a sharp degradation in zero-shot natural generalization. To resolve that, we design Generalization-Pivot Decoupling (GPD) by leveraging the difference in learning rate scheduling. This decouples the proxy transfer process into a generalization-anchored warm-up that maintains generalization and a generalization-pulled HPT that promotes adversarial robustness, to achieve an equilibrium between natural generalization and adversarial robustness. Extensive experiments on 15 zero-shot datasets demonstrate the effectiveness of our HPT-GPD method. The code is available at the website of github.com/fxw13/HPT-GPD.

\end{abstract}

\begin{document}

\section{Introduction}

Vision-Language Models (VLMs)~\cite{li2022align,li2022blip,singh2022revisiting} pre-trained on large-scale multi-modal datasets have achieved remarkable success in various practical applications, such as image classification~\cite{jia2021scaling,radford2021learning}, image caption generation~\cite{li2022blip,chen2022visualgpt}, medical assistance~\cite{liu2023clip,tang2022self}, \textit{etc}. Taking the foundational CLIP~\cite{radford2021learning} as an example, given a query image and a set of textual class labels, CLIP computes the similarity between the image embedding and the text embedding of each class, and predicts the class as the one with the highest similarity. Despite its strong zero-shot generalization in downstream tasks, VLMs are susceptible to adversarial perturbations designed to be human imperceptible~\cite{goodfellow2014explaining, madry2017towards,croce2020reliable,szegedy2013intriguing,moosavi2016deepfool}.

To address the threat posed by adversarial attacks, adversarial fine-tuning (AFT)~\cite{mao2022understanding, Wang_2024_CVPR} has been proposed to enhance the zero-shot adversarial robustness of large-scale models. However, existing methods often struggle to achieve satisfactory performance. In traditional classification tasks, adversarial robustness transfer~\cite{goldblum2020adversarially, chen2020robust, huang2023boosting, jung2024peeraid} has proven effective in improving adversarial robustness. These strategies typically involve first adversarially training a highly robust source model and then employing techniques such as knowledge distillation to transfer this robustness to a vulnerable target model, thereby enhancing its defensive capabilities. But it encounters significant challenges when applied to VLMs. Constructing an adversarial robust teacher  from scratch for large-scale multi-modal models incurs substantial computational costs, which becomes a considerable bottleneck for applying this paradigm in VLMs.

\begin{figure}[t]
	\centering
	\begin{minipage}{0.49\linewidth}
		\centering
		\includegraphics[height=0.71\linewidth, width=0.97\linewidth]{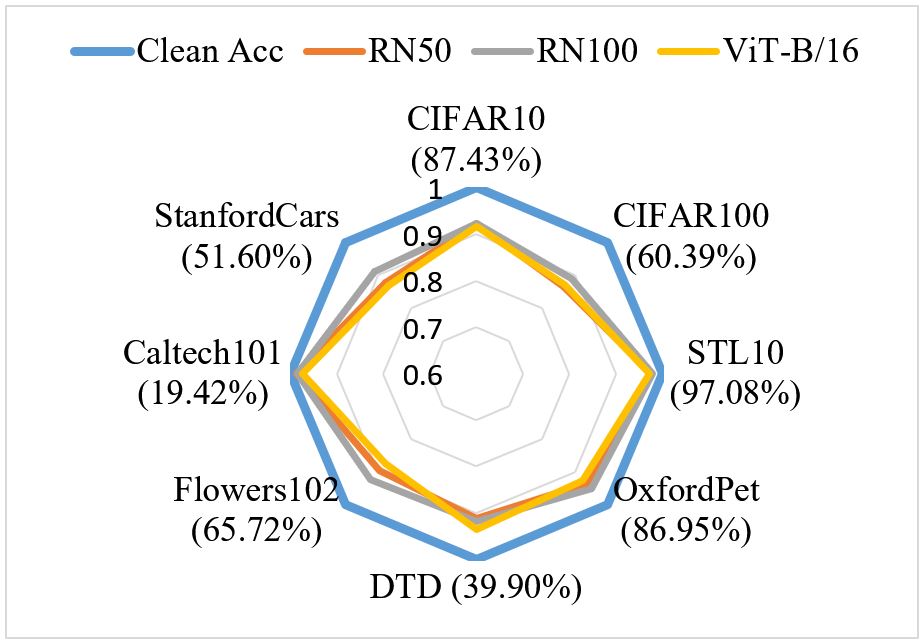}
        \caption*{{\scriptsize (a) ViT-B/32-based proxy.}}
		\label{chutian1}
	\end{minipage}
	\begin{minipage}{0.49\linewidth}
		\centering
		\includegraphics[width=0.97\linewidth]{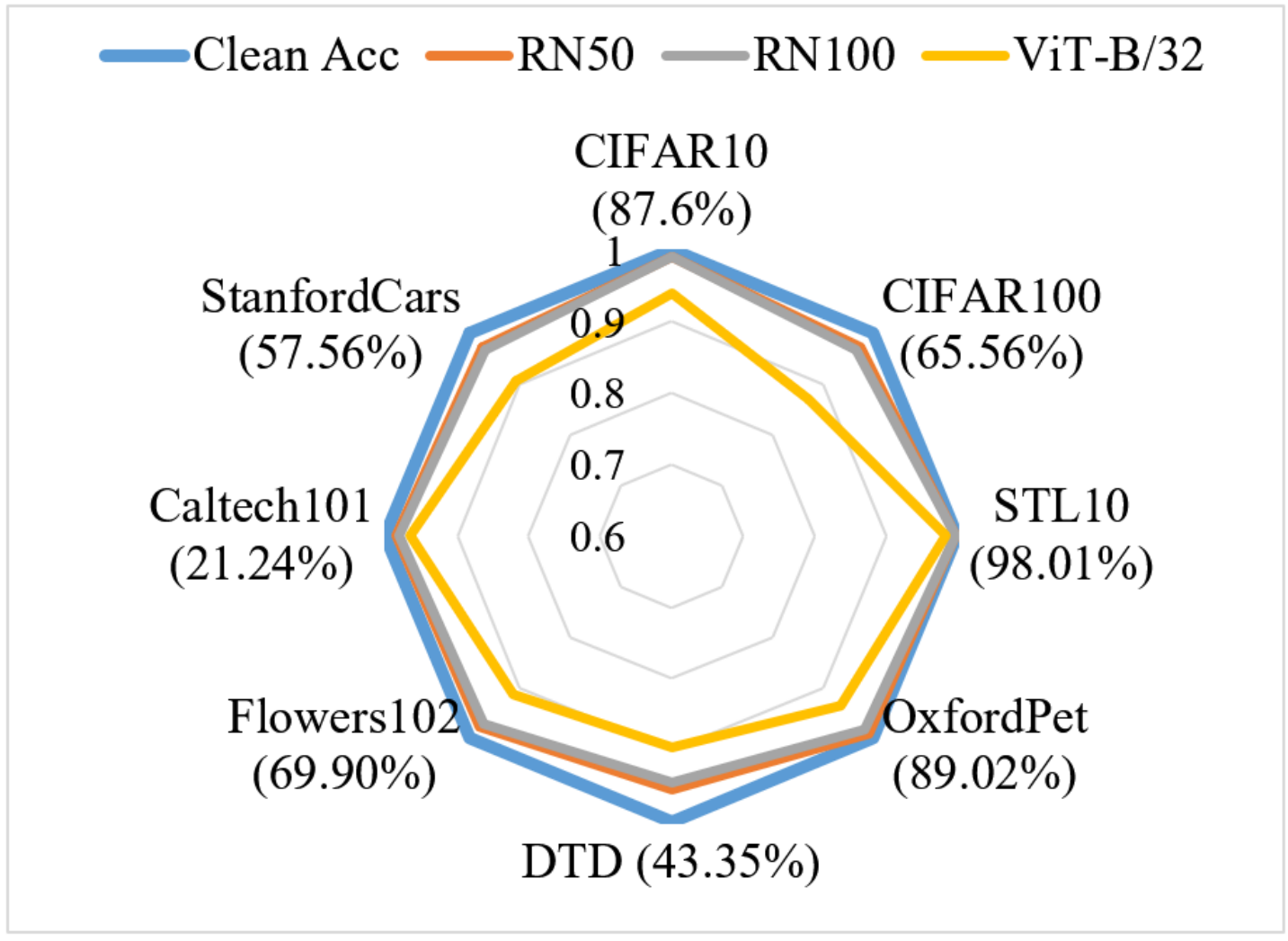}
		\caption*{{\scriptsize (b) ViT-B/16-based proxy.}}
		\label{chutian2}
	\end{minipage}
	
	\begin{minipage}{0.49\linewidth}
		\centering
		\includegraphics[width=0.97\linewidth]{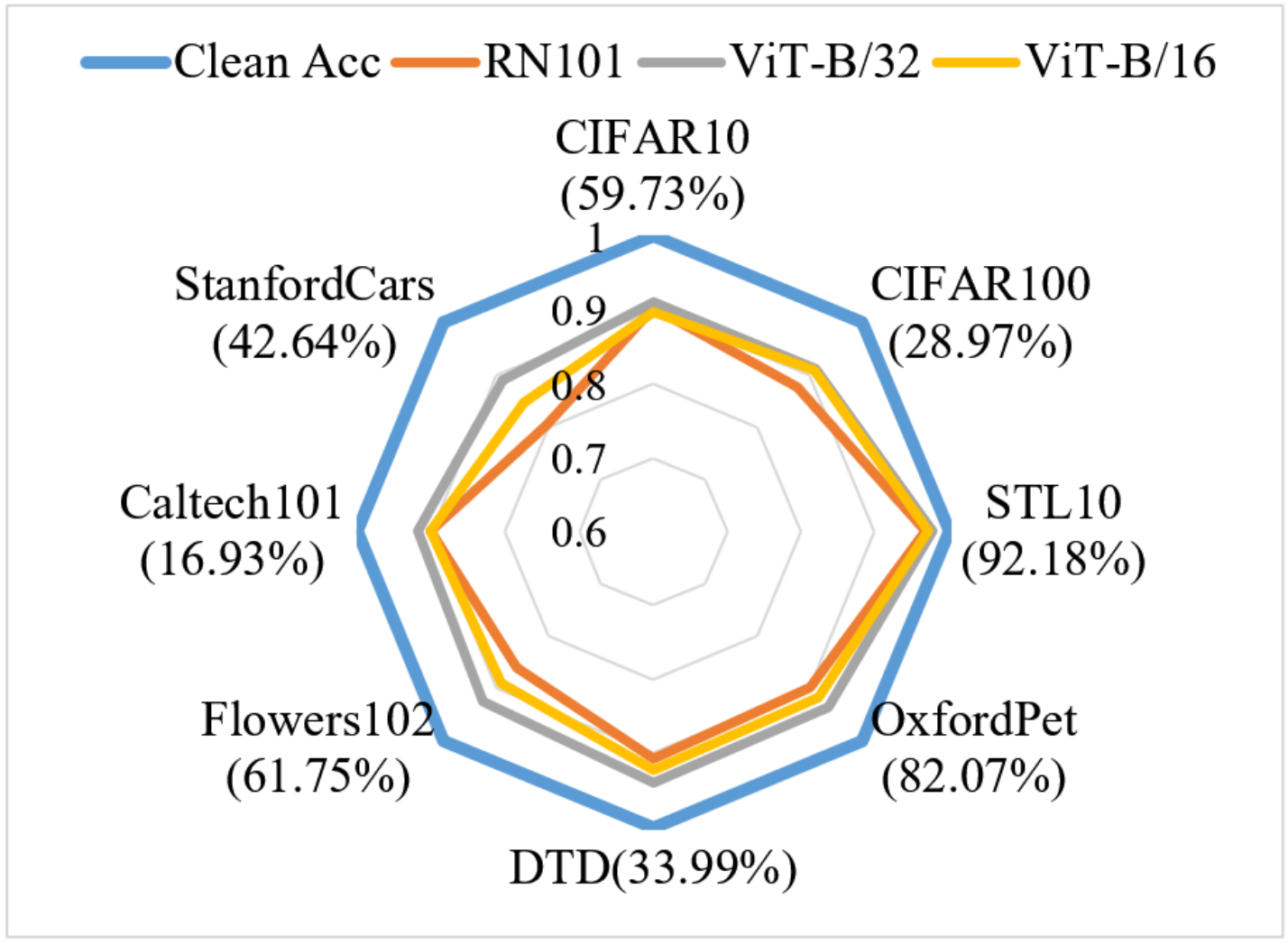}
        \caption*{{\scriptsize (c) ResNet50-based proxy.}}
		\label{chutian3}
	\end{minipage}
	\begin{minipage}{0.49\linewidth}
		\centering
		\includegraphics[width=0.97\linewidth]{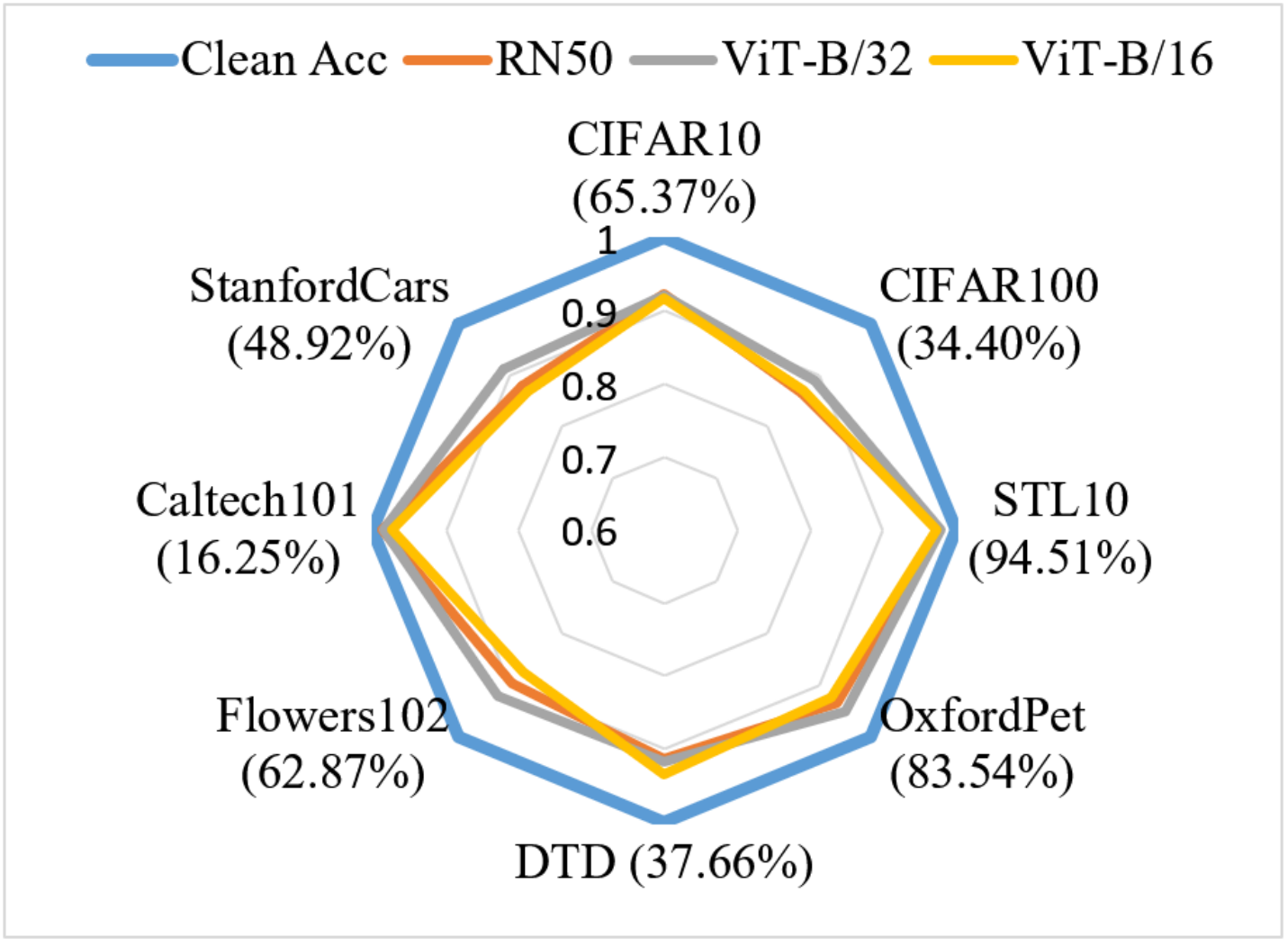}
		\caption*{{\scriptsize (d) ResNet101-based proxy.}}
		\label{chutian4}
	\end{minipage}
\caption{On 8 downstream datasets, variants of CLIP are used as proxies to evaluate the performance of adversarial samples generated by heterogeneous CLIPs (different colors). (a), (b), (c), and (d) are the performance of proxy CLIPs based on ViT-B/32, ViT-B/16, ResNet50 (RN50), and ResNet101 (RN101), respectively, tested on adversarial samples generated by other three heterogeneous CLIPs. The accuracy in brackets corresponds to the performance on clean samples.}
\label{fig1}
\end{figure}
We observe an intriguing phenomenon that can relieve this thorny problem: one CLIP model without adversarial training demonstrates inherent defensive capabilities against adversarial samples generated by its heterogeneous CLIP variants (i.e., models with different architectures). We define this phenomenon as \textbf{proxy adversarial robustness}. Specifically, CLIP has several image encoder variants, including ViT-B/16, ViT-B/32, ResNet50 (denoted as RN50) and ResNet101 (denoted as RN101). Taking these variants as examples, Fig.~\ref{fig1} illustrates the extent of the performance degradation in classification accuracy when these four CLIP variants serve as proxy adversarial defense providers, evaluated against adversarial samples generated by the other three variants on 8 downstream datasets. PGD-10 with $l_{\infty}$ norm perturbation bounds of 1 / 255 is used for testing. Fig.~\ref{fig1} (a), (b), (c), and (d) represent the performance of CLIPs based on ViT-B/32, ViT-B/16, RN50, and RN101 in providing proxy robustness, respectively. We have two key observations: Firstly, when faced with adversarial perturbations generated by heterogeneous CLIP models, the proxy CLIP models that do not undergo adversarial training maintain most of their performance. For instance, as shown in Fig.~\ref{fig1} (a), the ViT-B/32-based CLIP retains a performance degradation of no more than 20\% across multiple downstream datasets when evaluated against adversarial samples generated by other CLIPs. This demonstrates the existence of proxy adversarial robustness among heterogeneous CLIPs. Secondly, the strength of proxy adversarial defense is closely correlated with the architecture. For example, CLIP based on ViT significantly outperforms its ResNet-based counterpart. Thus, by transferring the adversarial robustness of the proxy CLIP model, it effortlessly enhances the zero-shot adversarial defense capabilities of the target CLIP model.

\begin{figure}[t]
\centering
\includegraphics[width=1.0\columnwidth, height=0.5\columnwidth]{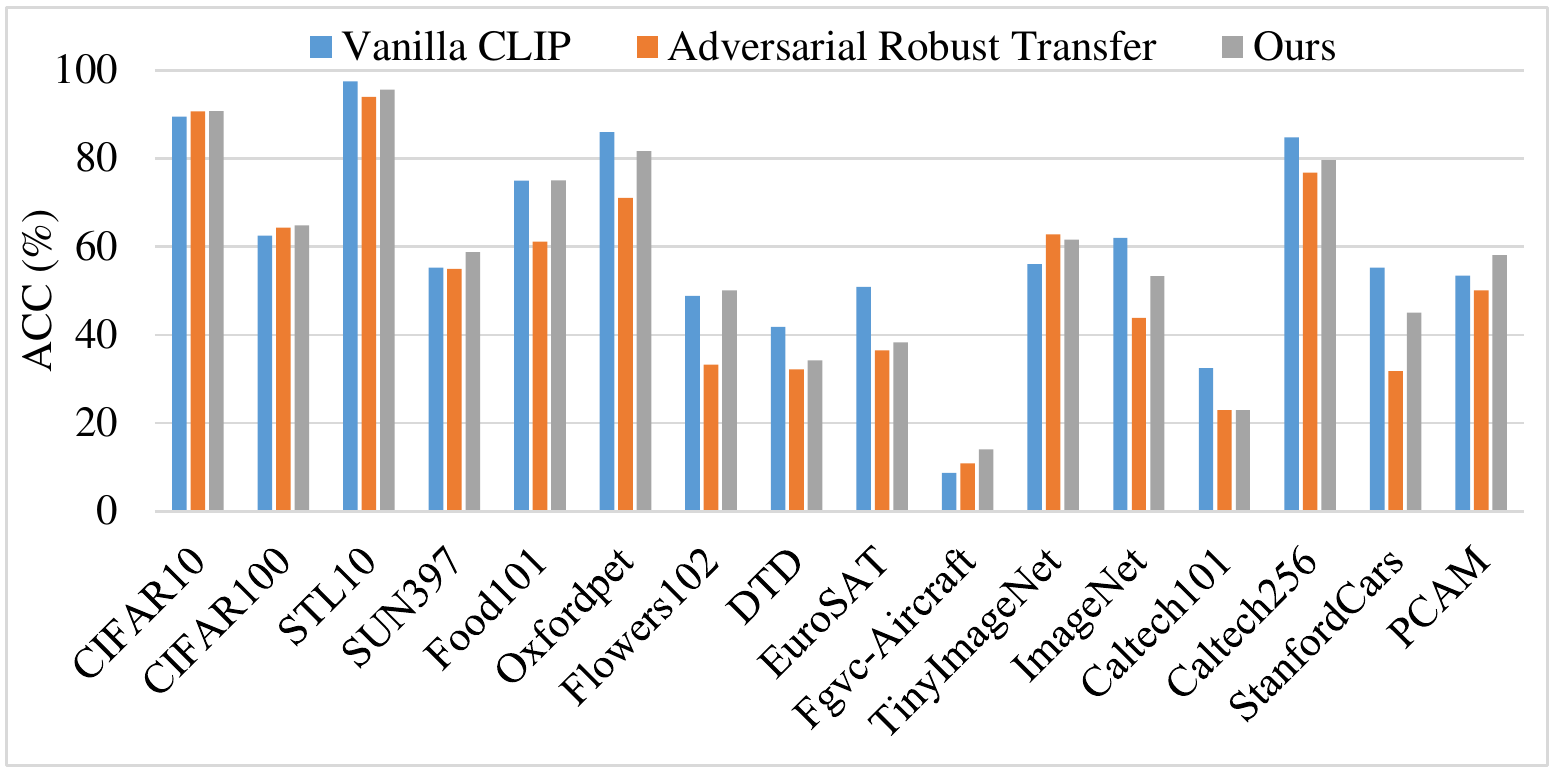}
\caption{When performing adversarial robustness transfer from ViT-B/16-based CLIP to ViT-B/32-based CLIP, the zero-shot generalization of the target CLIP is significantly impacted, leading to substantial degradation. After deploying our method, the performance on natural samples across downstream datasets is effectively preserved.}
\label{fig2}
\end{figure}

However, directly employing distillation as usual to transfer proxy adversarial robustness leads to severe overfitting to the dataset for adversarial fine-tuning. This results in substantial degradation of zero-shot natural generalization of the target CLIP, as illustrated in Fig.~\ref{fig2}. Specifically, after transferring proxy robustness from a ViT-B/16-based CLIP to a ViT-B/32-based CLIP via distillation pipeline, the ViT-B/32-based CLIP experiences a drastic degradation in natural robustness on multiple downstream datasets (e.g., from 75\% to 61\% on the Food101).

In traditional image classification tasks, researchers address the trade-off between adversarial and natural robustness by decoupling them during adversarial training~\cite{wang2023generalist}, which inspires us to propose a Heterogeneous Proxy Transfer framework via Generalization-Pivot Decoupling (HPT-GPD). It enhances the zero-shot adversarial robustness without compromising the natural generalization of the target CLIP through proxy transfer. Specifically, during adversarial fine-tuning, a generalization-anchored warm-up with a low learning rate is conducted to impart adversarial robustness to the target CLIP while maintaining the natural generalization ability of the vanilla CLIP. Then, the information of the optimized generalization network is injected to pull the HPT process which uses a higher learning rate to transfer proxy robustness via distillation. GPD ensures a balance between zero-shot adversarial robustness and the natural generalization of CLIP. The contributions of this paper are as follows:

\begin{itemize}
\item We reveal that vanilla CLIP can provide proxy adversarial robustness for heterogeneous CLIP models, establishing the foundation for the adversarial robust transfer of VLMs.
\item We propose a heterogeneous proxy transfer framework via generalization pivot decoupling, which effortlessly enhances the zero-shot adversarial robustness of the target CLIP while preserving its natural generalization ability. \item Experiments on 15 downstream datasets demonstrate that the proposed method offers significant improvements in both zero-shot adversarial robustness and natural generalization for CLIP, surpassing all other approaches.
\end{itemize}

\section{Related Work}
\subsection{Zero-shot Adversarial Robustness}

Earlier approaches~\cite{zhang2023atzsl,yucel2022robust} combining adversarial robustness~\cite{athalye2018obfuscated,carlini2017towards,dong2018boosting} with zero-shot learning~\cite{liu2019attribute,xian2018feature,jia2021scaling} relied on attributes, achieved only slight improvements, and generalized poorly without attribute annotations. There are also some works focusing on improving the generalization of adversarial robustness against unknown attacks~\cite{Zhang_MID, Fu_M3C,Tang_RobustOver,Fu_ASR}. In recent years, pre-trained large vision-language models (VLMs)~\cite{caron2021emerging,chen2022visualgpt,radford2021learning} have attracted much attention. They have strong zero-shot generalization capabilities and rely on multimodal information, eliminating the need for attributes. Some works focus on the zero-shot adversarial robustness of VLMs represented by CLIP.
FT-TeCoA~\cite{mao2022understanding} aligns the text embeddings and the adversarial visual features with contrastive adversarial fine-tuning. PMG~\cite{Wang_2024_CVPR} preserves the generalization features during adversarial fine-tuning by knowledge distillation from the original CLIP. LAAT~\cite{li2024language} increases the distance between text embeddings, but at the expense of losing the original semantic relevance, which reduces zero-shot generalization. However, these methods are not satisfactory in terms of natural generalization and zero-shot robust defense capabilities.

\subsection{Adversarial Robust Transfer}

Many researchers have studied adversarial robust transfer on traditional image classification tasks to provide the robustness of a larger teacher network to a student network via adversarial distillation. ARD~\cite{goldblum2020adversarially} adopts the concept of standard knowledge distillation, where the robust teacher network’s predictions on natural data guide the training of the student network. AKD\text{²}~\cite{chen2020robust} demonstrates that adversarial distillation, combined with weight averaging~\cite{hwang2021adversarial} to mitigate the robust overfitting problem. IAD~\cite{zhu2021reliable} highlights the diminishing reliability of the teacher network during adversarial distillation, it mitigates this issue by gradually shifting reliance from the teacher to the student’s self-derived knowledge as training progresses. RSLAD~\cite{zi2021revisiting} highlights the importance of utilizing robust soft labels during the inner optimization process to generate adversarial examples. AdaAD~\cite{huang2023boosting} demonstrates that maximizing the prediction discrepancy between a student network and a robust teacher network can improve the inner maximization process of adversarial training. PeerAiD~\cite{jung2024peeraid} trains a peer model from the student attacked sample to build a peer tutor with better guidance for adversarial distillation. However, these traditional robust transfer methods share a common prerequisite: \textit{the teacher network must undergo adversarial training}. This makes them difficult to apply for enhancing the zero-shot adversarial robustness of VLMs.

\section{Preliminaries}

In this paper, we investigate the zero-shot adversarial generalization capability of Vision-Language Models (VLMs) and use the basic image classification task to illustrate the proposed method. Taking the widely-used multi-modal CLIP as a representative example, let $F_{\theta}(\cdot)$ denote the image encoder of CLIP parameterized by $\theta$, and $G_{\phi}(\cdot)$ represent the text encoder parameterized by $\phi$. Given an input image $x$ and a corresponding category textual description $t$, such as ``This is a photo of $\{\}$'', where $\{\}$ contains the text of class, the image encoder $F_{\theta}(\cdot)$ and the text encoder $G_{\phi}(\cdot)$ output the corresponding deep feature embeddings $F_{\theta}(x)$ and $G_{\phi}(t)$, respectively. In the context of the image classification task, each category is associated with a specific text description, resulting in a total of $c$ categories. CLIP computes the similarity between the image feature embedding $F_{\theta}(x)$ and each category's text embedding $G_{\phi}(t_{m})$, where $t_{m}$ corresponds to the text description of the $m$-th category. The model outputs a $c$-dimensional vector representing the similarity scores, and the category with the highest similarity is selected as the final classification result: \begin{equation}
C\!(x,\!t)\! \!=\! \!\left[\operatorname{sim}\!\left(\!F\!_{\theta}(x), \! G\!_{\phi}\!\left(t_{1}\right)\right),\!  \ldots,\! \operatorname{sim}\!\left(\!F\!_{\theta}(x), \! G\!_{\phi}\!\left(t_{c}\right)\right)\right].
  \label{eq0}
\end{equation} 

\subsection{Adversarial Knowledge Distillation}

\begin{figure*}[t]
	\centering
    \includegraphics[height=0.38\linewidth, width=0.9\linewidth]{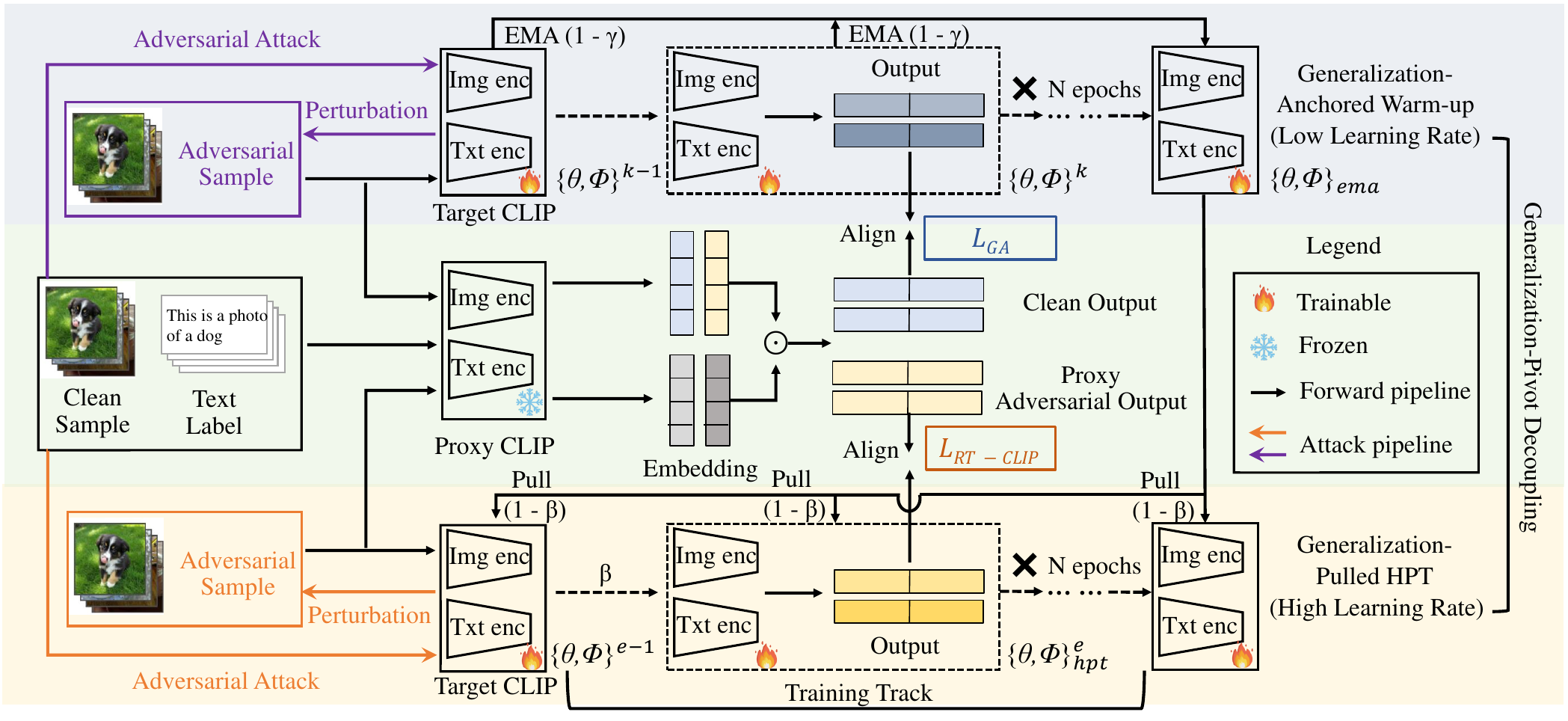}
	\caption{The pipeline of the proposed HPT-GPD. In the generalization-anchored warm-up, the zero-shot generalization ability is maintained at a low learning rate; then, the proxy robust transfer is performed at a high learning rate, while the model obtained in the warm-up is used for generalization pulling.}
	\label{fig3}
\end{figure*}

In traditional image classification tasks, knowledge distillation is typically performed for adversarial robust transfer. ARD~\cite{goldblum2020adversarially} provides a basic pipeline for adversarial knowledge distillation. Specifically, let $T(\cdot)$ and $S(\cdot)$ represent the teacher network and the student network, respectively, then the robust transfer loss is: \begin{equation}
L_{\mathrm{\emph{RT}}}\!=\!\mathbb{E}_{x \sim D}  {\left[\alpha \mathrm { KL } \left(S(x), T(x))\right.\right.}\!\!+\!\!(1-\alpha) CE\left(S(x), y\right)],
  \label{eq1}
\end{equation} where $\mathrm { KL }$ is the Kullback-Leibler divergence, $x$ is an input to the networks drawn from data generating distribution $D$, y is the corresponding ground-truth label, $\alpha$ is the modulation parameter usually set as 1. Most adversarial robust transfer works follow this basic paradigm and make improvements for different issues. However, they all require adversarial training for the teacher network, either offline or online. ARD also shows that by using traditional classification models, non-robust teachers will lead to non-robust students. Therefore, conventional adversarial robust transfer strategies are not suitable for the VLMs.
\subsection{Adversarial Fine-Tuning}
Adversarial Training (AT) is a typical strategy to promote defense by generating attack samples to update the model. However, deploying AT from scratch using the original training data of VLM requires expensive computational burden. Therefore, researchers focus on performing Adversarial Fine-Tuning (AFT)~\cite{mao2022understanding, Wang_2024_CVPR} on one downstream dataset to improve the zero-shot adversarial generalization. To clearly state the proposed method, we first introduce some basic notations: Let $D_d=\left \{ x,y,t \right \}_{i}^{N}$ denotes the downstream data that is for AFT, where $x_{i}$ is labeled by $y_{i}$, $t$ is the category text description, $N$ is the number of samples. Typically, AFT updates the model weights to solve a min-max optimization in a perspective of saddle point problem:\begin{equation}
\min _\theta \mathbb{E}_{(x_{i},y_{i},t)\sim D_d} \frac{1}{N} \sum_{i=1}^N \max _{\|\delta\|_p \leq \epsilon} {L}(C\left({x}_i+\delta, t), y_i \right),
  \label{eq2}
\end{equation} where ${x}_i+\delta$ is the adversarial sample generated during the AFT process that we denote as ${x}_i^{a}$ within the $\epsilon$-ball (bounded by the $L_p$ norm) centered at the natural input ${x}_i$, $\delta$ is the adversarial perturbation, $C(\cdot)$ is the output of CLIP in Eq.~\ref{eq0}. $L(\cdot)$ represents the supervision loss. AFT can be seen as a data augmentation strategy that supplements the training data with adversarial samples to adapt the model to the distribution of adversarial attacks, thereby improving adversarial robustness.

\section{Methodology}


We propose a Heterogeneous Proxy Transfer (HPT) strategy to build an information distillation channel between heterogeneous CLIPs to improve the zero-shot adversarial robustness of the target model. However, monotonous knowledge distillation will seriously damage the zero-shot generalization ability of the target CLIP for downstream natural samples. Therefore, we decouple HPT into generalization-anchored learning and generalization-pulled learning processes depending on the speed of the learning process, preserving the zero-shot generalization ability of the model through this Generalization-Pivot Decoupling (GPD), as shown in Fig.~\ref{fig3}. We expect the zero-shot adversarial robustness of the target CLIP to be significantly improved, while the generalization performance on downstream natural samples can be maintained by the proposed HPT-GPD.

\subsection{Heterogeneous Proxy Transfer}
\begin{figure}[t]
	\centering
    \includegraphics[height=0.4\linewidth, width=1.0\linewidth]{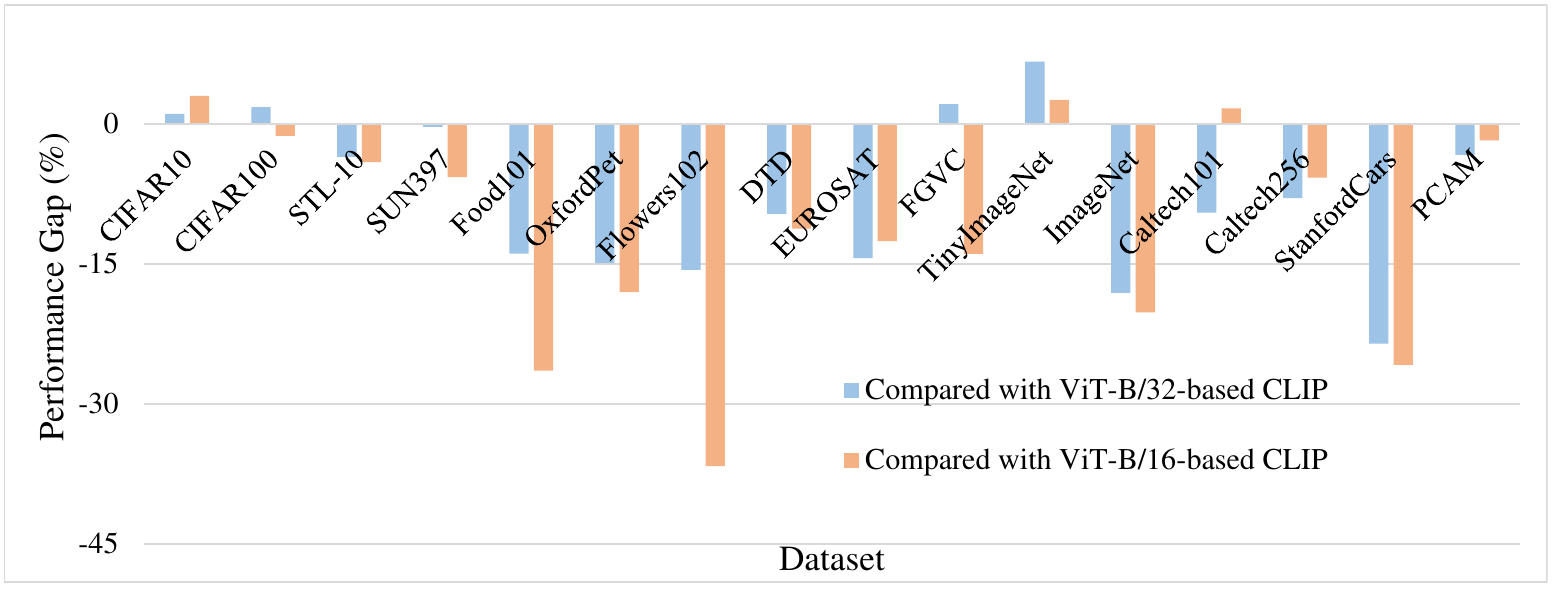}
	\caption{When using ViT-B/16-based CLIP as the proxy and ViT-B/32-based CLIP as the target model, the natural performance is reduced after the robust transfer via Eq.~\ref{eq4}.}
	\label{drop}
\end{figure}

We choose a vanilla CLIP as a proxy to provide proxy adversarial robustness, denoted as $P(\cdot)$. The target CLIP is a model that we expect to robustify in zero-shot adversarial situations, which is denoted as $T(\cdot)$. The probability distributions $\mathcal{P}_D$ of their classification outputs could be written as: \begin{equation} \begin{aligned}
\mathcal{P}_D({T(x_{i}^{a},t)})=\operatorname{softmax}(T(x_{i}^{a},t)),\\
\mathcal{P}_D({P(x_{i}^{a},t)})=\operatorname{softmax}(P(x_{i}^{a},t)),
\label{eq3}
\end{aligned}
\end{equation} where $t$ is the category textual descriptions, the outputs of $P(\cdot)$ and $T(\cdot)$ are generated similar to Eq.~\ref{eq0}, and the softmax function is used to normalize the magnitude. Most previous robust transfer strategies on traditional classification tasks are to generate a linkage pipeline between the teacher model and the student model, then optimize the parameters of the teacher and the student simultaneously during the adversarial training process. Different from them, we freeze the parameters of $P(\cdot)$ and use its predictions of adversarial examples generated by $T(\cdot)$ as soft labels for adversarial robust transfer:
\begin{equation}
L_{\mathrm{\emph{RT-CLIP}}}= \frac{1}{N} \sum_{i=1}^N { \mathrm { KL } (\mathcal{P}_D(T(x_{i}^{a},t))\| \mathcal{P}_D(P(x_{i}^{a},t)))},
\label{eq4}
\end{equation} where $x_{i}^{a}$ is the adversarial sample generated by the target CLIP model $T(\cdot)$, $\mathrm { KL }$ is the Kullback-Leibler divergence: \begin{equation}
K L(A \| B)=\sum_{j} A(j) \log \left(\frac{A(j)}{B(j)}\right).
\label{eq5}
\end{equation}
By transferring knowledge at the prediction distribution level, we expect the proxy adversarial robustness from $P(\cdot)$ could enhance the defense capability of the target model.


However, robust transfer using Eq.~\ref{eq4} will severely damage the zero-shot generalization. As shown in Fig.~\ref{drop}, when using ViT-B/16-based CLIP as the proxy and ViT-B/32-based CLIP as the target model, there is a sharp decrease in natural performance on downstream datasets. 

\begin{figure}[t]
\centering
\includegraphics[width=1.0\columnwidth, height=0.35\columnwidth]{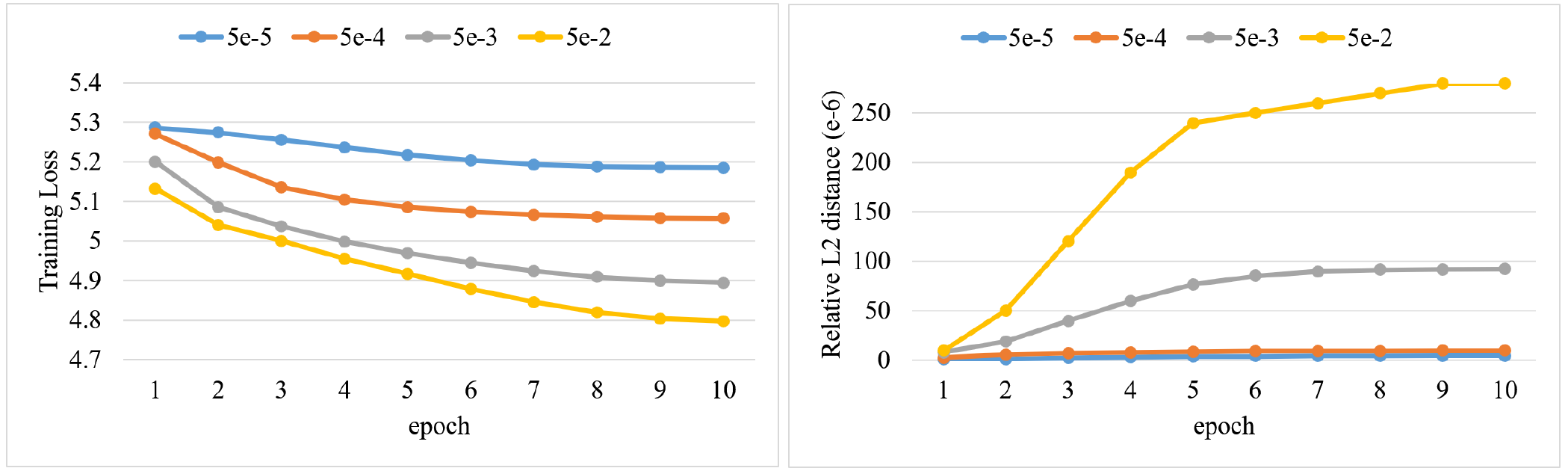}
\caption{Training loss (left) and relative $L_2$ distance of parameters between fine-tuned CLIP and original ViT-B/32-based CLIP (right) for AFT at different learning rates on the TinyImageNet dataset.}
\label{fig_loss}
\end{figure}

Generalist~\cite{wang2023generalist} has proved that adopting separate training strategies for clean and adversarial samples could balance their robustness. Fig.~\ref{fig_loss} shows the AFT dynamics of a ViT-B/32-based CLIP on the TinyImageNet under varying learning rates. The loss quantifies the model's adaptation efficacy to adversarial perturbations in downstream tasks, whereas the parameter distance captures the risk of catastrophic forgetting of natural generalization capabilities (as described in PMG~\cite{Wang_2024_CVPR}). Through a carefully regulated learning process, we demonstrate that optimizing adversarial robustness can coexist with preserving zero-shot generalization. The proposed Generalization-Pivot Decoupling (GPD) achieves an equilibrium between the two competing objectives.


\subsection{Generalization-Pivot Decoupling }
\textbf{Generalization-Anchored Warm-up. }We expect to preserve zero-shot generalization by decoupling the proxy robust transfer process. To this end, we propose to warm up the target CLIP with a low learning rate, and use the predictions of the proxy model $P(\cdot)$ on natural samples as the soft labels for pre-trained model guiding. Trades~\cite{zhang2019theoretically} proves that aligning the predicted distributions of natural and adversarial samples is beneficial to the balance between natural and adversarial performance. The probability distribution of prediction for clean samples with $P(\cdot)$ is:
\begin{equation}
\begin{aligned}
\mathcal{P}_D({P(x_{i},t)})=\operatorname{softmax}(P(x_{i},t)).
\label{eq6}
\end{aligned}
\end{equation}
In proxy adversarial transfer, we usually choose a more powerful proxy CLIP as the teacher model, which may have a stronger zero-shot generalization ability than the target CLIP. Thus, guidance from a pre-trained model during fine-tuning is beneficial. Based on this, the generalization-anchored loss $L_{\mathrm{\emph{GA}}}$ is: \begin{equation}
\begin{aligned}
L_{\mathrm{\emph{GA}}}= \frac{1}{N} \!\sum_{i=1}^N \!{ \mathrm { KL } (\mathcal{P}_D(T(x_{i}^{a},t))\| \mathcal{P}_D(P(x_{i},t)))}.
  \label{eq7}
\end{aligned}
\end{equation} By optimizing Eq.~\ref{eq7}, we expect to give the target model adversarial robustness initially while maintaining natural generalization ability.

In addition, in the generalization-anchored warm-up, vanilla CLIP usually has the strongest zero-shot generalization ability. The target model in the early stage of training may also provide meaningful generalization information. Therefore, it is necessary to integrate the knowledge of the entire training trajectory to maintain zero-shot generalization ability. To this end, we use the Exponential Moving Average (EMA) strategy to preserve the model in the training sequence: \begin{equation}
\left \{ \theta ,\phi \right  \}_{ema}^{k} \!\!=\!\! \left\{
\begin{array}{cc}  
  \! \!\!\gamma \! \cdot \! \left \{ \theta ,\phi \right  \}^{k-1}_{ema}\!+\!(1-\gamma) \! \cdot \!
\left \{ \theta ,\phi \right  \}^{k} & \!\!\!\text{if } k > 0 ,\\[2mm]
 \!\!\! \left \{ \theta ,\phi \right  \}_{vanilla} & \!\!\!\text{if } k = 0,
\end{array}
\right.
 \label{eq8}
\end{equation} where $\left \{ \theta ,\phi \right  \}_{ema}^{k}$ represents the image encoder and text encoder parameters maintained recursively by EMA at the $k$-th epoch, $\left \{ \theta ,\phi \right  \}^{k}$ is the model parameters generated after the current epoch update, and $\left \{ \theta ,\phi \right  \}_{vanilla}$ is the pre-trained CLIP parameters. $\gamma$ is the decay factor and is set as 0.9 as usual. Ultimately, through generalization-anchored warm-up learning, the model parameters generated by EMA will have strong zero-shot generalization and initial adversarial robustness.


\textbf{Generalization-Pulled HPT. }We expect to further improve the adversarial robustness of the target CLIP through robust transfer. Specifically, we optimize Eq.~\ref{eq4} with a high learning rate. Besides, a generalization-pulled strategy is introduced in HPT to inject the zero-shot generalization information into the optimization process periodically. Specifically, the EMA model obtained during the above generalization-anchored warm-up provides generalization information, forcing the HPT training process not to deviate too far from the generalization space: \begin{equation}
\left \{ \theta ,\phi \right  \}_{hpt}^{e} \!\!=\!\! \left\{
\begin{array}{cc}  
  \! \!\!\beta  \! \cdot \! \left \{ \theta ,\phi \right  \}^{e-1}\!+\!(1-\beta ) \! \cdot \!
\left \{ \theta ,\phi \right  \}_{ema} & \!\!\!\text{if } e > 0 ,\\[2mm]
 \!\!\! \left \{ \theta ,\phi \right  \}_{vanilla} & \!\!\!\text{if } e = 0,
\end{array}
\right.
 \label{eq9}
\end{equation} where $\left \{ \theta ,\phi \right  \}_{hpt}^{e}$ is the initialization network parameter at the beginning of $e$-th epoch during HPT, $\left \{ \theta ,\phi \right  \}_{ema}$ is the EMA results of generalization-anchored learning, $\beta$ is the trade-off parameter and we set it as 0.5. Because we believe that the zero-shot generalization and adversarial robustness of a model are equally crucial.

The process of generalization pulling is to linearly inject the model parameters of the original CLIP with strong generalization into the model generated by proxy robust transfer in each epoch. In this way, although the model needs to change the parameter distribution to adapt to adversarial samples constantly, the periodic information injection will prevent it from deviating too far (too much damage to natural generalization). Ultimately, we expect $T(\cdot)$ to have strong zero-shot natural generalization and adversarial robustness, and the performance is evaluated on it.

\textbf{Overall optimization.} As mentioned before, the training of $T(\cdot)$ is decoupled into two stages: during the generalization-anchored warm-up, the zero-shot generalization ability is maintained through the supervision of Eq.~\ref{eq7} at a low learning rate; then, Eq.~\ref{eq4} is optimized at a high learning rate to perform the proxy robust transfer, while the model obtained in the previous step is used for generalization pulling. Through Heterogeneous Proxy Transfer via Generalization-Pivot Decoupling, the target CLIP achieves a balance of adversarial robustness and natural generalization.

\subsection{Theoretical Analysis}

We provide a theoretical analysis of the proposed generalized anchored loss $L_{GA}$ and demonstrate the overall optimization of Eq.~\ref{eq2}. This explains why our method is able to maintain natural accuracy while continuously improving adversarial robustness, as shown in Table~\ref{tabel2}.
\subsubsection{Theoretical analysis of $L_{GA}$ Loss in Eq.~\ref{eq7}.}

\begin{theorem}\label{bound}
Let $T$ be the target model and $P$ be the proxy model, then the total expected risk $\epsilon(T)$ of $T$ on both adversarial samples denoted by $x^{a}$ and clean samples denoted by $x$ is bounded by:
\begin{equation}
\begin{aligned}
\epsilon(T)
&= \epsilon_{\text{adv}}(T) + \epsilon_{\text{cln}}(T) \\
&\leq
\underbrace{\epsilon_{\text{adv}}(T, P)}_{\text{Attack Defense}}
+ \underbrace{\epsilon_{\text{cln}}(T, P)}_{\text{Clean Maintenance}}
+ \underbrace{2\epsilon_{\text{cln}}(P)}_{\text{Proxy Error}},
\end{aligned}
\label{eq12}
\end{equation}
where:
\begin{itemize}
    \item $\epsilon_{\text{adv}}(T) = \mathbb{E}[L(T(x^a), y)]$: risk of $T$ on adversarial samples.
    \item $\epsilon_{\text{cln}}(T) = \mathbb{E}[L(T(x), y)]$: risk of $T$ on clean samples.
    \item $\epsilon_{\text{adv}}(T, P) = \mathbb{E}[L(T(x^a), P(x))]$: discrepancy between $T$ on adversarial samples and $P$ on clean samples.
    \item $\epsilon_{\text{cln}}(T, P) = \mathbb{E}[L(T(x), P(x))]$: discrepancy between $T$ and $P$ on clean samples.
    \item $\epsilon_{\text{cln}}(P) = \mathbb{E}[L(P(x), y)]$: risk of $P$ on clean samples.
    \item $L(\cdot,\cdot)$: loss function measuring the divergence between predictive distributions (e.g., cross-entropy loss and KL divergence).
\end{itemize}

\end{theorem}

\noindent \textit{Proof.}
Assuming that $L(\cdot,\cdot)$ satisfies the triangle inequality
$L(A,C) \le L(A,B)+L(B,C)$ in expectation (which holds for convex losses such as cross-entropy or KL divergence), for the risk of adversarial samples, there is:
\begin{equation}\label{adv_risk}
\begin{aligned}
\epsilon_{\text{adv}}(T) &= \mathbb{E}[L(T(x^a), y)] \\
&\leq \mathbb{E}[L(T(x^a), P(x))] + \mathbb{E}[L(P(x), y)] \\
&= \epsilon_{\text{adv}}(T,P) + \epsilon_{\text{cln}}(P).
\end{aligned}
\end{equation}

\noindent Similarly, for the risk of clean samples, there is:
\begin{equation}\label{cln_risk}
\begin{aligned}
\epsilon_{\text{cln}}(T) &= \mathbb{E}[L(T(x), y)] \\
&\leq \mathbb{E}[L(T(x), P(x))] + \mathbb{E}[L(P(x), y)] \\
&= \epsilon_{\text{cln}}(T,P) + \epsilon_{\text{cln}}(P).
\end{aligned}
\end{equation}

\noindent Proof is done, by combining Eq.\ref{adv_risk} with Eq.\ref{cln_risk}.

Our method minimizes the bound in Theorem \ref{bound}. We analyze each term of Eq.~\ref{eq12} as follows:

\begin{itemize}
\item Attack Defense: $L_{GA}$ minimizes $\epsilon_{\text{adv}}(T, P)$ by aligning the predictions on adversarial samples of $T$ with the predictions on clean samples of $P$ to improve adversarial robustness.

\item Clean Maintenance: The EMA strategy preserves the generalization capability of the target model by maintaining parameters close to the original CLIP, implicitly controlling $\epsilon_{\text{cln}}(T, P)$.

\item Proxy Error: The $\epsilon_{\text{cln}}(P)$ is small since we use high-performing CLIP as a proxy, which exhibits strong zero-shot generalization.
\end{itemize}

Overall, Theorem~\ref{bound} theoretically justifies that minimizing $L_{GA}$ reduces the discrepancy between adversarial and clean risks, providing a formal explanation for the robustness improvements observed in experiments. Similarly, the theory for Eq.~\ref{eq4} is shown in Theorem \ref{theo_2}.
\begin{theorem}\label{theo_2}
Let $T$ be the target model and $P$ be the proxy model, then the total expected risk $\epsilon(T)$ of $T$ on both adversarial samples denoted by $x^{a}$ and clean samples denoted by $x$ is bounded by:
\begin{equation}
\begin{aligned}
\epsilon(T)
&= \epsilon_{\text{adv}}(T) + \epsilon_{\text{cln}}(T) \\
&\leq
\underbrace{\epsilon_{\text{adv}}(T, P)}_{\text{Attack Defense}}
+ \underbrace{\epsilon_{\text{cln}}(T, P)}_{\text{Clean Maintenance}}
+ \underbrace{\epsilon_{\text{adv}}(P)+\epsilon_{\text{cln}}(P)}_{\text{Proxy Error}},
\end{aligned}
\label{eq13}
\end{equation}
where $\epsilon_{\text{adv}}(T, P) = \mathbb{E}[L(T(x^a), P(x^a))]$, and other terms are completely the same as Eq.\ref{eq12}. The proof and analysis of Eq.~\ref{eq4} are similar to Theorem 1, thus we omit it.
\end{theorem}

\subsubsection{Theoretical analysis of Adversarial Fine-tuning Loss in Eq.~\ref{eq2}.}
In the study of adversarial defense, the min-max optimization objective has been well established for adversarial training, theoretically grounded in the concept of \emph{adversarial risk minimization} by Madry et al.~\cite{madry2017towards}. This defines the adversarial risk as a saddle point problem, where the inner maximization approximates the worst-case adversary, and the outer minimization finds robust model parameters. Besides, this formulation shares a similar optimization structure with that in Generative Adversarial Networks (GANs)~\cite{goodfellow2014gan}, where a saddle point problem is solved to reach a Nash equilibrium between the generator and discriminator by a min-max procedure.

By analogy, in Eq.~\ref{eq2}, the inner maximization acts as an ``adversary generator'', seeking the most aggressive perturbation to fool the CLIP model, while the outer minimization training corresponds to the ``defender'', updating the CLIP parameters to be robust against above attacks. However, directly optimizing this objective in VLMs leads to severe overfitting and natural accuracy degradation, as shown in Fig.~\ref{fig2}. Our two-stage strategy addresses this by decoupling the min-max optimization: the generalization-anchored warm-up establishes a robust initialization while preserving natural generalization, and then the generalization-pulled HPT performs adversarial enhancement from this favorable starting point. This two-phase strategy 
is empirically validated through comprehensive experiments on 15 datasets. This demonstrates successful minimization of the adversarial risk while maintaining natural generalization accuracy, as evidenced in Table~\ref{tabel2}.

\section{Experiments}
\begin{table*}[t]
\centering
\resizebox{\textwidth}{!}
{\begin{tabular}{clcccccccccccccccc>{\columncolor{gray!10}}cc}
\hline
\multirow[c]{1}{*}{Metrics}
& \multirow[c]{1}{*}{Method}
& \multicolumn{1}{c}{\rotatebox{45}{\footnotesize CIFAR10}} & \multicolumn{1}{c}{\rotatebox{45}{\footnotesize CIFAR100}} & \multicolumn{1}{c}{\rotatebox{45}{\footnotesize STL-10}}& \multicolumn{1}{c}{\rotatebox{45}{\footnotesize SUN397}}& \multicolumn{1}{c}{\rotatebox{45}{\footnotesize Food101}}  & \multicolumn{1}{c}{\rotatebox{45}{\footnotesize OxfordPet}} & \multicolumn{1}{c}{\rotatebox{45}{\footnotesize Flower102}}& \multicolumn{1}{c}{\rotatebox{45}{\footnotesize DTD}} & \multicolumn{1}{c}{\rotatebox{45}{\footnotesize EUROSAT}}   & \multicolumn{1}{c}{\rotatebox{45}{\footnotesize FGVC}} & \multicolumn{1}{c}{\rotatebox{45}{\footnotesize TinyImageNet}}& \multicolumn{1}{c}{\rotatebox{45}{\footnotesize ImageNet}} & \multicolumn{1}{c}{\rotatebox{45}{\footnotesize Caltech101}} & \multicolumn{1}{c}{\rotatebox{45}{\footnotesize Caltech256}} & \multicolumn{1}{c}{\rotatebox{45}{\footnotesize StanfordCars}}  & \multicolumn{1}{c}{\rotatebox{45}{\footnotesize PCAM}}  & \multicolumn{1}{c}{\small \textbf{Average}}& \multicolumn{1}{c}{\small \textbf{Time (s)}}
\\ \hline
\multirow{5}{*}{Adv acc}  & CLIP  &42.18 &17.57 &67.77 &10.49 &3.28 &8.98 &4.88 &18.75 &3.84 &0.39 &0.39 &6.09 &15.82 &23.76 &2.34 &51.61 &17.38 &0\\
\cline{2-20}
& FT-Standard  &24.21 &12.30 &47.65 &5.71 &2.65 &1.56 &7.61 &17.57 &0.13 &0.00 &0.39 &6.87 &\textbf{20.31} &23.82 &1.75 &48.15 &13.79 &406\\
& FT-TeCoA~\cite{mao2022understanding}  &40.82 &24.41 &70.70 &19.21 &14.45 &28.13 &23.05 &28.13 &12.57 &3.13& 19.33& 16.48& 24.02 &40.56& 12.69 &53.68 &26.96 &1112\\
& PMG~\cite{Wang_2024_CVPR} &67.06	&33.31	&77.91	&11.13	&13.91	&42.00	&27.58	&29.63	&21.25	&3.30	&15.15	&9.99	&14.82 &41.21	&13.32 	&57.65 &29.95 &1817\\
& HPT-GPD (ours) &\textbf{71.14}	&\textbf{40.10}	&\textbf{83.88}	&\textbf{24.56}	&\textbf{22.24}	&\textbf{50.45}	&\textbf{28.04}	&\textbf{32.34}	&\textbf{25.69}	&\textbf{3.69}	&\textbf{19.62}	&\textbf{18.53}	&19.11 &\textbf{50.06}	&\textbf{14.90}	&\textbf{58.15}	&\textbf{35.16} &1486\\
\hline
\hline
\multirow{5}{*}{Clean acc}& CLIP   &87.43	&60.39	&97.08 &57.79 &82.93	&86.95	&65.72	&39.90	&41.12	&20.50	&56.06 &58.24 &19.42 &79.43	&51.60	& 53.52 &59.88 &0
\\
\cline{2-20}
& FT-Standard  &88.67 &58.78 &95.89 &46.61 &70.07 &75.00 &41.60 &41.79 &38.60 &8.57 &58.59 &51.28 &29.10 &79.88 &37.10 &54.74 &54.76 &406\\
& FT-TeCoA~\cite{mao2022understanding}  &66.79 &41.01 &89.25 &47.01 &52.81 &70.31 &36.13 &35.94 &18.88 &7.81 &48.83 &43.67 &28.32 &72.98 &37.89 &37.89 &46.99 &1112\\
& PMG~\cite{Wang_2024_CVPR} &82.80	&52.28	&92.99	&58.78	&70.52	&84.46	&56.77	&35.11	&28.58	&15.75	&46.41	&54.86 	&20.98	&76.82	&48.04 &	57.70	&55.18 &1817\\
& HPT-GPD (ours) &90.74	&64.83	&95.60	&58.79	&75.05	&81.68&	50.11	&34.26 &38.32 	&14.07 	&61.56	&53.35	&22.87 &79.63	&45.00	&58.14	&57.75 &1486\\ \hline
\end{tabular}}
\caption{Average zero-shot adversarial accuracy (i.e., Adv acc) and clean accuracy (i.e., Clean acc) under PGD-10 attack. The target ViT-B/32-based CLIP is fine-tuned on TinyImageNet and then evaluated on 15 downstream datasets, proxy CLIP is ViT-B/16 based. The time cost for each epoch of different methods is provided. \textbf{Bold} indicates the best results.}
\label{tabel1}
\end{table*}

\begin{table*}[htbp]
\centering
\resizebox{\textwidth}{!}
{\begin{tabular}{clcccccccccccccccc>{\columncolor{gray!10}}c}
\hline
& \multirow[c]{1}{*}{Method}
& \multicolumn{1}{c}{\rotatebox{45}{\footnotesize CIFAR10}} & \multicolumn{1}{c}{\rotatebox{45}{\footnotesize CIFAR100}} & \multicolumn{1}{c}{\rotatebox{45}{\footnotesize STL-10}}& \multicolumn{1}{c}{\rotatebox{45}{\footnotesize SUN397}}& \multicolumn{1}{c}{\rotatebox{45}{\footnotesize Food101}}  & \multicolumn{1}{c}{\rotatebox{45}{\footnotesize OxfordPet}} & \multicolumn{1}{c}{\rotatebox{45}{\footnotesize Flower102}}& \multicolumn{1}{c}{\rotatebox{45}{\footnotesize DTD}} & \multicolumn{1}{c}{\rotatebox{45}{\footnotesize EUROSAT}}   & \multicolumn{1}{c}{\rotatebox{45}{\footnotesize FGVC}} & \multicolumn{1}{c}{\rotatebox{45}{\footnotesize TinyImageNet}}& \multicolumn{1}{c}{\rotatebox{45}{\footnotesize ImageNet}} & \multicolumn{1}{c}{\rotatebox{45}{\footnotesize Caltech101}} & \multicolumn{1}{c}{\rotatebox{45}{\footnotesize Caltech256}} & \multicolumn{1}{c}{\rotatebox{45}{\footnotesize StanfordCars}}  & \multicolumn{1}{c}{\rotatebox{45}{\footnotesize PCAM}}  & \multicolumn{1}{c}{\small \textbf{Average}}
\\ \hline
& CLIP   &8.98 &3.90 &13.47 &0.07 &0.39 &0.00 &0.00 &0.00 &0.00 &0.12 &0.00& 0.00 &0.41 &0.15& 0.09 &0.16 &1.74\\
& PMG~\cite{Wang_2024_CVPR} &25.96	&11.72	&40.11	&0.47	&0.44	&0.63	&0.54	&1.01	&0.03	&0.15	&0.19	&1.00	&0.20	&1.54	&\textbf{5.18}	&0.06	&5.58
\\
& HPT-GPD (ours) &\textbf{34.00}	&\textbf{18.16}	&\textbf{59.13}	&\textbf{4.55}	&\textbf{5.12}	&\textbf{10.06}	&\textbf{4.75}	&\textbf{5.53}	&\textbf{0.12}	&\textbf{0.33}	&\textbf{2.32}	&\textbf{7.08}	&\textbf{10.21}	&\textbf{19.52}	&1.20	&\textbf{0.87}	&\textbf{11.43}
 \\ \hline
\end{tabular}}
\caption{Average zero-shot robust accuracy (\%) and clean accuracy (\%) under strong AutoAttack. The performance is compared with PMG. \textbf{Bold} indicates the best results.}
\label{tabel2}
\end{table*}

\subsection{Experimental Setup}

\textbf{Datasets}. Following the settings in PMG~\cite{Wang_2024_CVPR}, we evaluate the zero-shot generalization and robustness of the CLIP model fine-tuned on the TinyImgeNet~\cite{deng2009imagenet} dataset. We report the performance of the models on the TinyImgeNet test set and 15 zero-shot downstream task test data, which include: CIFAR10~\cite{krizhevsky2009learning}, CIFAR100~\cite{krizhevsky2009learning},
STL10~\cite{coates2011analysis}, ImageNet~\cite{deng2009imagenet}, Caltech101~\cite{fei2006one}, and Caltech256~\cite{griffin2007caltech} for general object recognition task; scene recognition represented by SUN397~\cite{xiao2010sun}; OxfordPets~\cite{parkhi2012cats}, Flowers102~\cite{nilsback2008automated}, FGVCAircraft~\cite{maji2013fine}, and StanfordCars~\cite{krause20133d} for fine-grained recognition; Food101~\cite{bossard2014food}, EuroSAT~\cite{helber2019eurosat}, and DTD~\cite{cimpoi2014describing} for domain-specific task; PCAM~\cite{bejnordi2017diagnostic} for medical image recognition. To meet the input requirements of the CLIP model, the images in these datasets are preprocessed to a size of 3 $\times$ 224 $\times$ 224 using bicubic interpolation
followed by channel-wise normalization with mean and standard deviation computed by ImageNet.

\textbf{Baselines}. Considering that zero-shot adversarial robustness research is still in its infancy and the number of available methods is limited, our main results focus is on comparing our method with the current SOTA method FT-TeCoA~\cite{mao2022understanding} and PMG~\cite{Wang_2024_CVPR}. The specific implementation of the proposed method is based on PMG.

\textbf{Implementation Details}. Current works FT-TeCoA and PMG use CLIP based on ViT-B/32 as an example to verify the robust performance. Following the same setting, unless otherwise specified, we use CLIP based on ViT-B/32 as the target model and select CLIP based on ViT-B/16 as a strong proxy. For fairness, all baselines are trained and evaluated under the same hyper-parameter settings, including learning rate, number of epochs, attack budget, and so on. All datasets share the same hyperparameter configuration. We perform adversarial fine-tuning on the target model for 10 epochs using the SGD~\cite{robbins1951stochastic} optimizer with a batch size of 64. For the generalization-anchored warm-up, we set the base learning rate to 5e-5. For generalization-pulled HPT, we use the base learning rate of 5e-2. PGD-2 and PGD-10 with $l_{\infty}$ norm perturbation bounds of 1 / 255 are used for training and testing, respectively. Besides, to verify the proxy adversarial robustness and the versatility of the proposed method, we use ResNet-based CLIP and the ALIGN~\cite{jia2021scaling} model which is different from CLIP to verify the cross-architecture performance. The model is trained on one NVIDIA GeForce RTX 3090 GPU.

\subsection{Main Result}

\textbf{Performance Comparison.} We fine-tune the ViT-B/32-based CLIP with the ViT-B/16 CLIP as a proxy on TinyImageNet and then evaluate it on all 16 downstream datasets. It is worth noting that on datasets other than TinyImageNet, the evaluation is performed in the zero-shot manner. During training and evaluation, we use the PGD-10 attack with a perturbation bound $\epsilon= 1/255$. The accuracy results are shown in Table~\ref{tabel1}. As can be seen, compared with the SOTA PMG, our method improves robust accuracy by an average of 5.21\% and achieves improvements on most datasets. Besides, our method outperforms PMG in terms of clean accuracy and is slightly higher than the accuracy of fine-tuning on a clean dataset (i.e., FT-Standard). Overall, our approach improves zero-shot adversarial robustness while maintaining natural generalization.

\textbf{Computational burden.} Since we introduced the proxy, the computational overhead has increased compared to FT-TeCoA, but it is much lower than PMG, and each epoch of training is 331 seconds faster. The trade-off between performance and computational cost makes HPT-GPD a highly practical solution. While FT-TeCoA is lighter and PMG is heavier, our approach strikes a better balance, making it the recommended choice for most applications that require strong robustness without excessive cost. In resource-critical scenarios, FT-TeCoA remains a viable baseline; however, for scenarios where robustness is crucial, the significant performance improvement offered by HPT-GPD may be more important than computational cost.

\begin{table}[t]
\centering
\resizebox{0.35\textwidth}{!}{\begin{tabular}{cc|ccc}
\hline
\multicolumn{2}{c|}{Method}
& \multirow{2}{*}{Adv acc} & \multirow{2}{*}{Clean acc} & \multirow{2}{*}{Average}
\\ \cline{1-2}
$L_{RT-CLIP}$  & $L_{GA}$ & &  \multicolumn{1}{l}{}
\\ \hline
  $\times$ &$\times$&29.95 &55.18	&42.57\\
  \checkmark &$\times$  &\textbf{38.06} 	&52.31 	&45.19  \\
  \checkmark & \checkmark   & 35.16 &\textbf{57.75} &\textbf{46.46} \\
\hline
\end{tabular}}
\caption{Ablation analysis of different loss items under PGD-10 attack. The best results are shown in \textbf{bold}.}
\label{table3}
\end{table}

\subsection{Performance against AutoAttack}
AutoAttack~\cite{croce2020reliable} is a stronger attack strategy and is often deployed to verify the robustness of the model comprehensively. We use the standard version of AutoAttack for evaluation and give robust accuracy with a perturbation bound $\epsilon$ of 1/255 in Table~\ref{tabel2}. Specifically, following the settings of PMG, we use two variants of PGD, APGD-CE and APGD-DLR, for evaluation. AutoAttack shows a stronger attack and the robustness of the original CLIP model drops sharply. Our method also experiences a certain degree of decline, but it still improves the adversarial accuracy of the model compared to PMG.
\subsection{Ablation Study}

\begin{figure}[t]
	\centering
    \includegraphics[width=1.0\linewidth]{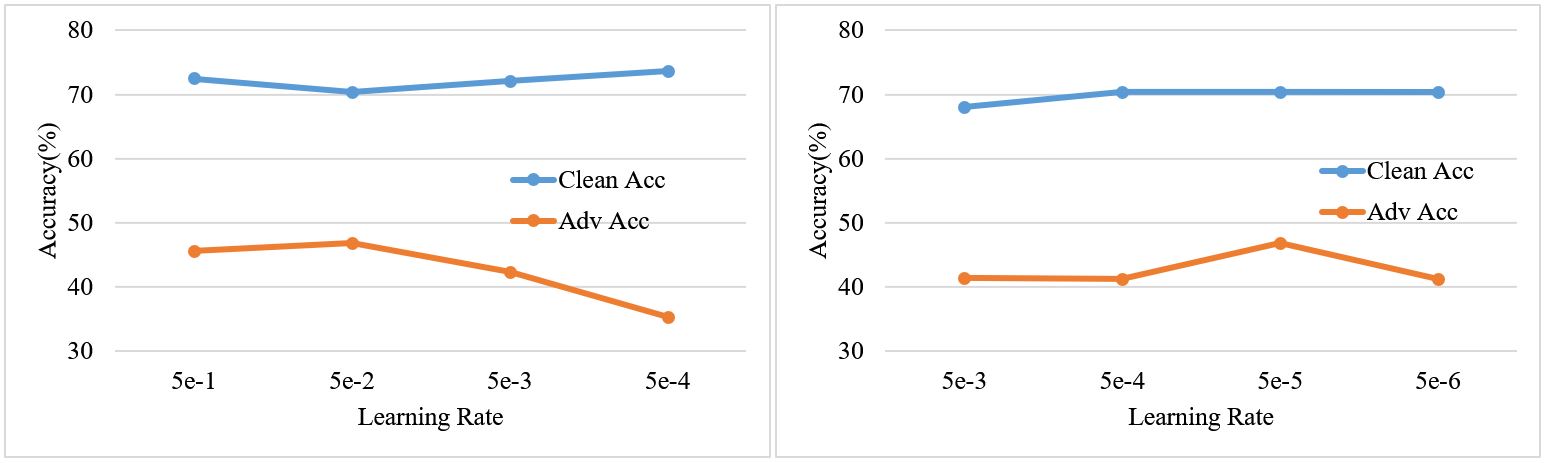}
	\caption{The influence of learning rate. The figures on the left and right represent the learning rates in the Generalization-Pulled HPT and Generalization-Anchored Warm-up stages, respectively.}
	\label{fig6}
\end{figure}

\begin{table*}[htbp]
\centering
\resizebox{\textwidth}{!}
{\begin{tabular}{clcccccccccccccccc>{\columncolor{gray!10}}c}
\hline
& \multirow[c]{1}{*}{Method}
& \multicolumn{1}{c}{\rotatebox{0}{\footnotesize CIFAR10}} & \multicolumn{1}{c}{\rotatebox{0}{\footnotesize CIFAR100}} & \multicolumn{1}{c}{\rotatebox{0}{\footnotesize STL-10}}& \multicolumn{1}{c}{\rotatebox{0}{\footnotesize SUN397}}& \multicolumn{1}{c}{\rotatebox{0}{\footnotesize Food101}}  & \multicolumn{1}{c}{\rotatebox{0}{\footnotesize OxfordPet}} & \multicolumn{1}{c}{\rotatebox{0}{\footnotesize Flower102}}& \multicolumn{1}{c}{\rotatebox{0}{\footnotesize DTD}} & \multicolumn{1}{c}{\rotatebox{0}{\footnotesize EUROSAT}}   & \multicolumn{1}{c}{\rotatebox{0}{\footnotesize FGVC}} & \multicolumn{1}{c}{\rotatebox{0}{\footnotesize Tiny}}& \multicolumn{1}{c}{\rotatebox{0}{\footnotesize ImageNet}} & \multicolumn{1}{c}{\rotatebox{0}{\footnotesize Caltech101}} & \multicolumn{1}{c}{\rotatebox{0}{\footnotesize Caltech256}} & \multicolumn{1}{c}{\rotatebox{0}{\footnotesize StanfordCars}}  & \multicolumn{1}{c}{\rotatebox{0}{\footnotesize PCAM}}  & \multicolumn{1}{c}{\small \textbf{Average}}
\\ \hline
& Clean Dis   &25.83&8.99&49.78&4.95&2.52&11.80&5.33&5.69&0.12&0.18&0.71&6.25&3.50&16.45&2.55&42.93&11.72

\\
& Ensemble &48.65&23.44&70.00&5.14&7.32&28.26&16.95&16.12&0.75&0.96&9.38&9.50&15.88&25.83&4.12&6.81&18.07
\\

& Adaptive Attack &65.98&40.77&84.08&38.88&48.67&67.48&39.06&29.31&22.78&8.67&27.42&35.94&20.51&63.99&32.19&53.90&42.48
\\
\hline
\end{tabular}}
\captionof{table}{Ablation analysis of distillation using clean samples (ViT-B/16-based to ViT-B/32-based CLIP, Clean Dis), ensemble of ViT-B/16-based and ViT-B/32-based CLIP in a non-adversarial setting (Ensemble), and Adaptive Attack for our method. Adv accuracy is reported.}
\label{tabel_dis}
\end{table*}
\begin{table*}[htbp]
\centering
\resizebox{\textwidth}{!}
{\begin{tabular}{clcccccccccccccccc>{\columncolor{gray!10}}c}
\hline
\multirow[c]{1}{*}{Metrics}
& \multirow[c]{1}{*}{Method}
& \multicolumn{1}{c}{\rotatebox{0}{\footnotesize CIFAR10}} & \multicolumn{1}{c}{\rotatebox{0}{\footnotesize CIFAR100}} & \multicolumn{1}{c}{\rotatebox{0}{\footnotesize STL-10}}& \multicolumn{1}{c}{\rotatebox{0}{\footnotesize SUN397}}& \multicolumn{1}{c}{\rotatebox{0}{\footnotesize Food101}}  & \multicolumn{1}{c}{\rotatebox{0}{\footnotesize OxfordPet}} & \multicolumn{1}{c}{\rotatebox{0}{\footnotesize Flower102}}& \multicolumn{1}{c}{\rotatebox{0}{\footnotesize DTD}} & \multicolumn{1}{c}{\rotatebox{0}{\footnotesize EUROSAT}}   & \multicolumn{1}{c}{\rotatebox{0}{\footnotesize FGVC}} & \multicolumn{1}{c}{\rotatebox{0}{\footnotesize Tiny}}& \multicolumn{1}{c}{\rotatebox{0}{\footnotesize ImageNet}} & \multicolumn{1}{c}{\rotatebox{0}{\footnotesize Caltech101}} & \multicolumn{1}{c}{\rotatebox{0}{\footnotesize Caltech256}} & \multicolumn{1}{c}{\rotatebox{0}{\footnotesize StanfordCars}}  & \multicolumn{1}{c}{\rotatebox{0}{\footnotesize PCAM}}  & \multicolumn{1}{c}{\small \textbf{Average}}
\\ \hline
\multirow{4}{*}{Adv acc}
& TeCoA-4/255  &11.73&9.00&30.00&1.94&1.19&3.22&3.38&4.79&10.62&0.12&\textbf{7.70}&\textbf{3.86}&4.82&10.02&\textbf{0.21}&23.55&7.89
 \\
& TeCoA-4/255+ours  &\textbf{45.20}&\textbf{14.71}&\textbf{64.75}&\textbf{5.90}&\textbf{2.63}&\textbf{3.79}&\textbf{6.64}&\textbf{27.66}&\textbf{12.21}&0.12&5.83&3.82&\textbf{14.61}&\textbf{21.53}&0.15&\textbf{54.67}&\textbf{17.76}
 \\ \cline{2-18}
& TeCoA-8/255  &2.30&1.70&4.28&0.02&0.04&0.03&0.07&1.22&4.94&0.00&0.17&0.20&1.18&1.54&0.00&3.88&1.35
 \\
& TeCoA-8/255+ours  &\textbf{11.51}&\textbf{2.48}&\textbf{28.78}&\textbf{3.14}&\textbf{0.17}&\textbf{0.22}&\textbf{0.98}&\textbf{13.94}&\textbf{5.49}&0.00&\textbf{0.93}&\textbf{2.06}&\textbf{8.54}&\textbf{7.71}&\textbf{0.01}&\textbf{5.77}&\textbf{5.73}

 \\
\hline
\hline
\multirow{4}{*}{Clean acc}
& TeCoA-4/255  &64.68&37.83&81.79&28.12&21.36&54.65&27.52&21.97&19.67&4.05&\textbf{65.36}&29.56&18.54&50.36&14.29&49.96&36.86
 \\
& TeCoA-4/255+ours  &\textbf{88.76}&\textbf{63.52}&\textbf{95.06}&\textbf{54.66}&\textbf{74.27}&\textbf{81.82}&\textbf{50.61}&\textbf{34.95}&\textbf{35.36}&\textbf{13.08}&57.61&\textbf{45.44}&\textbf{22.62}&\textbf{78.92}&\textbf{43.84}&\textbf{55.40}&\textbf{55.99 }
 \\ \cline{2-18}
& TeCoA-8/255  &49.43&21.39&71.85&28.38&16.36&46.85&21.86&15.37&17.29&6.15&28.35&24.46&16.93&44.03&14.08&\textbf{59.38}&30.14
 \\
& TeCoA-8/255+ours  &\textbf{85.68}&\textbf{59.49}&\textbf{94.08}&\textbf{55.12}&\textbf{62.44}&\textbf{73.04}&\textbf{40.64}&\textbf{31.65}&\textbf{27.22}&\textbf{9.42}&\textbf{49.99}&\textbf{45.04}&\textbf{21.63}&\textbf{74.79}&\textbf{34.67}&53.58&\textbf{51.15 }\\ \hline
\end{tabular}}
\caption{Zero-shot adversarial and clean accuracy under 4/255 and 8/255 perturbation budgets. \textbf{Bold} indicates the best results.}
\label{tabel3}
\end{table*}

\textbf{Contribution of loss items. }We perform ablation analysis on the two loss terms included in the proposed method, as shown in Table~\ref{table3}. As mentioned earlier, when robust transfer is conducted during adversarial fine-tuning, it will lead to severe overfitting on the downstream dataset. After the supervision of $L_{RT-CLIP}$, although the adversarial robustness is improved, the average natural generalization is reduced by 2.87\%. After deploying $L_{GA}$, the natural generalization is improved, and the average accuracy is also increased despite the drop in adversarial robustness. This shows that the proposed method can better balance zero-shot adversarial robustness and natural generalization.

\textbf{Influence of learning rate. }In the process of decoupled training, the choice of learning rate plays an important role. We fix the learning rate in one stage while changing the other to evaluate the performance. As shown in Fig.~\ref{fig6}, in the Generalization-Pulled HPT stage, a high learning rate is crucial for proxy robustness. It provides the model with the necessary momentum to escape the flat minimum and explore the parameter region that is resistant to adversarial perturbations. Low learning rate fails to promote this transition, resulting in insufficient robustness gain. In contrast, in the Generalization-Anchored Warm-up, a low learning rate is crucial to maintain the zero-shot generalization ability of the pre-trained model. It guides the optimization toward a flat minimum in the loss landscape, leading to improved generalization performance. Using a high learning rate will severely perturb the well-generalized weights, resulting in a significant drop in clean accuracy. This decoupling approach in learning rate during training provides a simple yet effective alternative to architectural decoupling methods.

\begin{figure}[t]
	\centering
	\begin{minipage}{0.49\linewidth}
		\centering
		\includegraphics[height=0.6\linewidth, width=0.77\linewidth]{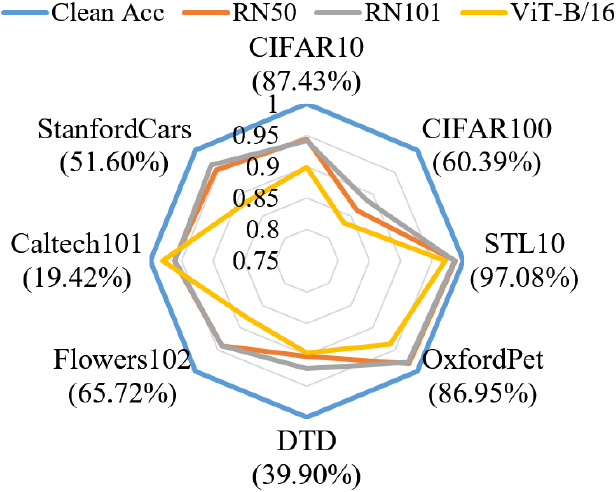}
		\label{chutian1}
	\end{minipage}
	\begin{minipage}{0.49\linewidth}
		\centering
		\includegraphics[height=0.6\linewidth, width=0.77\linewidth]{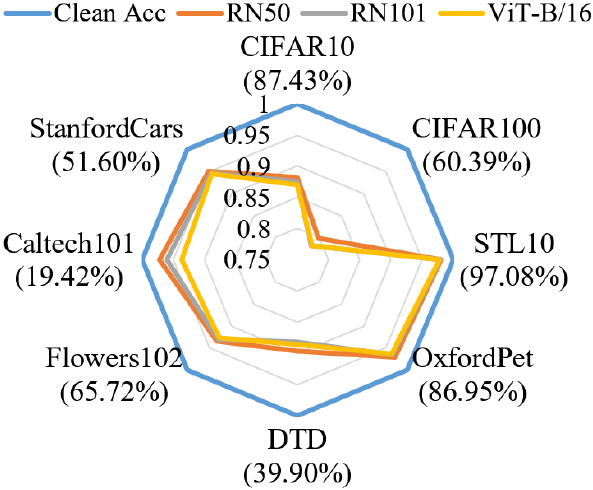}
		\label{chutian2}
	\end{minipage}
\caption{On 8 downstream datasets, different CLIPs are used as proxies to evaluate the adversarial samples generated by ViT-B/32-based CLIP. Left and right denote 4/255 and 8/255 perturbation budgets, respectively.}
\label{fig_budget}
\end{figure}

\textbf{Distillation strategy and
adaptive attack. }Table~\ref{tabel_dis} shows the defense results of distillation using only clean samples (Clean Dis) and a standard ensemble of proxy and target models in a non-adversarial setting (Ensemble). These two no-AT settings can not provide reliable robustness. The model will be fragile if it is not distilled adversarially. Since our strategy involves a proxy model, it is crucial to evaluate the robustness of the proposed method under adaptive attacks. During the attack, the perturbations are based on PGD, but the generation no longer depends solely on the target model; instead, it obtains information from the gradients of both models. As can be seen, the proposed method is robust against adaptive attacks that simultaneously perturb the target and proxy models. The main reason is that the target model naturally provides proxy robustness for the proxy model.

\subsection{Performance under Large Attack Budgets}

To comprehensively verify the effectiveness of the proposed method, we analyze its performance under adversarial perturbations of larger budgets. Specifically, we evaluate the results on downstream datasets using PGD-10 attacks with perturbations of 4/255 and 8/255. The models were adversarially fine-tuned and evaluated at the corresponding attack intensities. The results are shown in Table~\ref{tabel3}. The average robust accuracy of TeCoA continues to decrease with the increase in attack strength. However, our method consistently outperforms TeCoA at all attack levels. Besides, we evaluate proxy robustness under $l_{\infty}$ perturbations with budgets of 4/255 and 8/255: Using ViT-B/32-based CLIP as the target model, we generate adversarial examples via PGD-10 attacks and evaluate these examples through other CLIPs as proxies. As shown in Fig.~\ref{fig_budget} under large attack perturbations, the proxy maintains high robustness under increased budgets (at least 75\% of the clean accuracy in brackets). This demonstrates that the proxy can provide reliable robustness even under large attack strengths. These two figures together emphasize our motivation: the proxy model exhibits adversarial robustness against attacks produced by heterogeneous models, which can be effectively exploited in our framework.

\subsection{Extended Architectures}
\begin{figure}[t]
	\centering
    \includegraphics[width=1.0\linewidth]{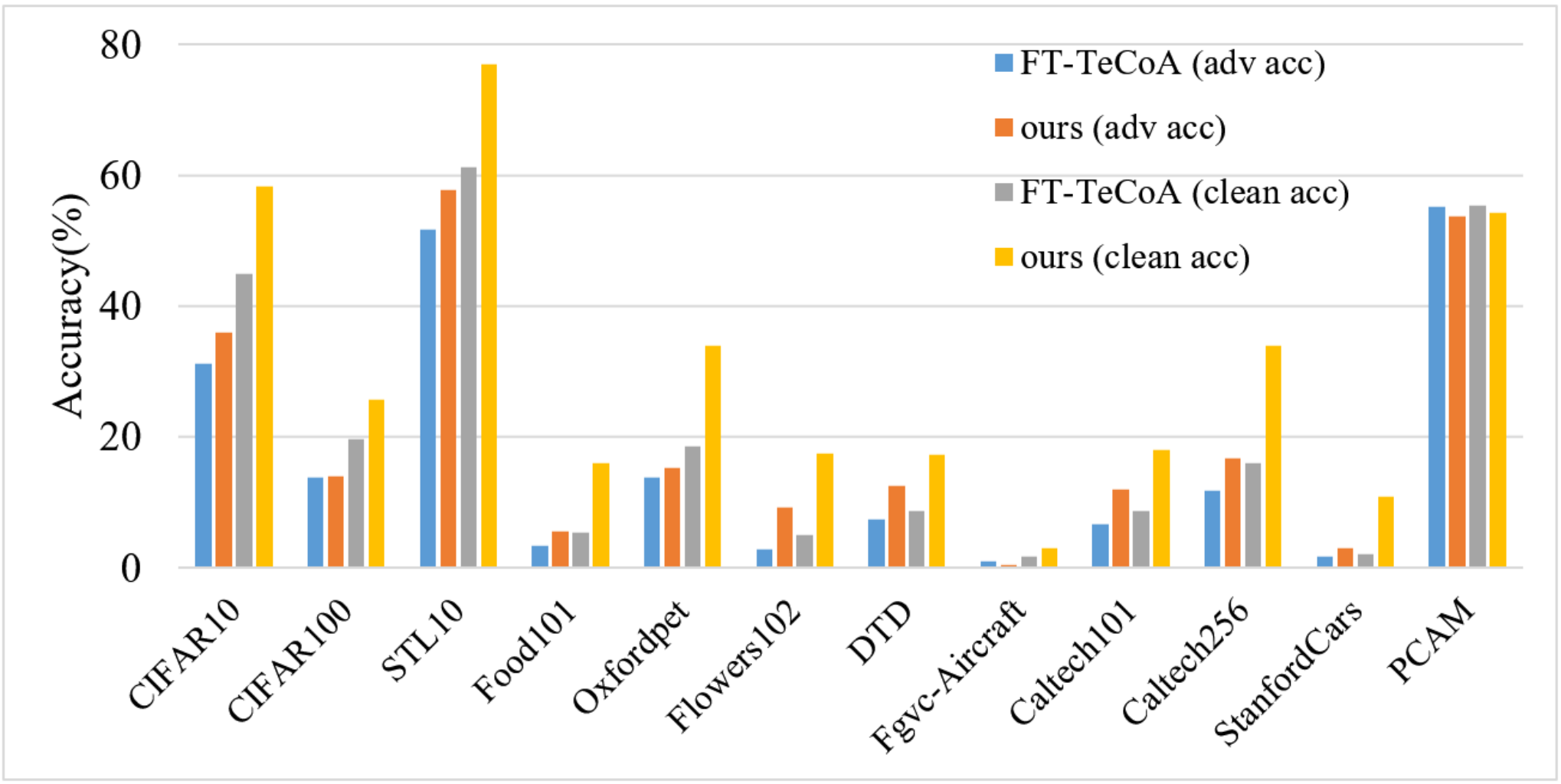}
	\caption{The proxy adversarial transfer performance from CLIP based on ViT-B/32 to CLIP based on ResNet50.}
	\label{fig5}
\end{figure}

\begin{table}[t]
\setlength{\tabcolsep}{2pt}
\centering
\resizebox{0.48\textwidth}{!}
{\begin{tabular}{clccccccccccc}
\hline
\multirow[c]{1}{*}{Metrics}
& \multirow[c]{1}{*}{Proxy}
& \multicolumn{1}{c}{\rotatebox{60}{\footnotesize Tiny}} & \multicolumn{1}{c}{\rotatebox{60}{\footnotesize CIFAR10}} & \multicolumn{1}{c}{\rotatebox{60}{\footnotesize CIFAR100}} & \multicolumn{1}{c}{\rotatebox{60}{\footnotesize STL-10}}& \multicolumn{1}{c}{\rotatebox{60}{\footnotesize Food101}}  & \multicolumn{1}{c}{\rotatebox{60}{\footnotesize OxfordPet}} & \multicolumn{1}{c}{\rotatebox{60}{\footnotesize Flower102}}& \multicolumn{1}{c}{\rotatebox{60}{\footnotesize DTD}} & \multicolumn{1}{c}{\rotatebox{60}{\footnotesize EUROSAT}}   & \multicolumn{1}{c}{\rotatebox{60}{\footnotesize FGVC}}
\\ \hline
\multirow{2}{*}{Adv acc}& CLIP   &31.56 &76.87	&45.33	&85.15	&28.30	&55.85	&28.23	&31.06&	31.33	&6.51
\\
& AFT   &51.88 &53.92& 27.36	&72.09	&9.65&	28.40 &	6.93	&13.72&	16.06&	2.04
\\ \hline
\multirow{2}{*}{Clean acc}& CLIP  &62.73 &90.65	&64.33	&93.96	&61.14	&71.03	&33.23	&32.18	&36.49	&10.86
 \\
& AFT &62.67	&68.11	&35.62	&82.20	&12.80&	34.15&	8.47&	14.68&	16.99 & 2.55

 \\ \hline
\end{tabular}}
\caption{The effect of whether the proxy has been adversarially fine-tuned on direct robust transfer. AFT causes severe overfitting during the transfer precess.}
\label{tabel4}
\end{table}

\textbf{Other CLIP.} We verify the proxy adversarial transfer on more CLIP architectures. Specifically, we use CLIP based on ViT-B/32 as the proxy and CLIP based on ResNet50 (RN50) as the target model to evaluate the proposed method. The results are shown in Fig.~\ref{fig5}. On most downstream datasets, the proposed method is significantly better than FT-TeCoA in zero-shot adversarial robustness and natural generalization. This proves the universality of the proposed method.

\textbf{AFT Proxy.} We evaluate the impact of whether the proxy model is adversarially fine-tuned before the robust transfer, as shown in Table~\ref{tabel4}. Specifically, the proxy ViT-B/16-based CLIP is adversarially fine-tuned on TinyImageNet, and the ViT-B/32-based CLIP is used as the target model of direct transfer. Compared with using vanilla CLIP as a proxy, CLIP after AFT will lead to more serious overfitting in robust transfer: although the accuracy on TinyImageNet is improved, the performance on multiple downstream datasets is drastically reduced. Therefore, it is inappropriate to pre-adversarially fine-tune the teacher model on the downstream dataset.

\textbf{Beyond CLIP.} ALIGN~\cite{jia2021scaling} is a multi-modal model that is completely different from CLIP. In Table~\ref{tabel5}, we verify the recognition performance of ALIGN on adversarial samples generated by CLIP, proving the existence of proxy adversarial robustness when the architectures are quite different. Besides, we evaluate the robust transfer performance when ALIGN is used as the proxy and CLIP as the target model. Robust transfer endows CLIP with a certain defense capability, and the final adversarial robustness is closely related to the proxy itself. Since the proposed method preserves the properties of the target model, the final performance is likely to surpass that of the proxy (clean accuracy on Tiny).

\begin{table}[t]
\centering
\resizebox{0.48\textwidth}{!}
{\begin{tabular}{cccccccccc}
\hline
\multirow[c]{1}{*}{Metrics}
& \multirow[c]{1}{*}{Model}
&  \multicolumn{1}{c}{\rotatebox{60}{\footnotesize Tiny}}	&\multicolumn{1}{c}{\rotatebox{60}{\footnotesize Food101}}	&\multicolumn{1}{c}{\rotatebox{60}{\footnotesize OxfordPet}}	&\multicolumn{1}{c}{\rotatebox{60}{\footnotesize Flowers102}}&	\multicolumn{1}{c}{\rotatebox{60}{\footnotesize DTD}} &\multicolumn{1}{c}{\rotatebox{60}{\footnotesize Caltech256}}&	\multicolumn{1}{c}{\rotatebox{60}{\footnotesize StanfordCars}}
\\ \hline
\multirow{2}{*}{Clean acc} & ALIGN~\cite{jia2021scaling}   &31.93	&67.58&	77.62&52.48&49.26&81.96	&44.57
\\
& HPT-GPD (ours) &51.39 &55.02	&65.22	&41.58	&30.32& 71.94	&26.12
\\
\hline
\multirow{2}{*}{Adv acc} & ALIGN~\cite{jia2021scaling} &28.89	&63.61	&75.88&	51.24&	49.10 &80.65	&43.64
 \\
& HPT-GPD (ours) &26.64	&15.01	&39.71	&21.00	&24.04 &41.14	&8.66
 \\
\hline
\end{tabular}}
\caption{Proxy adversarial robustness and robust transfer from ALIGN to ViT-B/32-based CLIP. }
\label{tabel5}
\end{table}

\begin{figure}[t]
	\centering
\includegraphics[height=0.23\linewidth, width=0.8\linewidth]{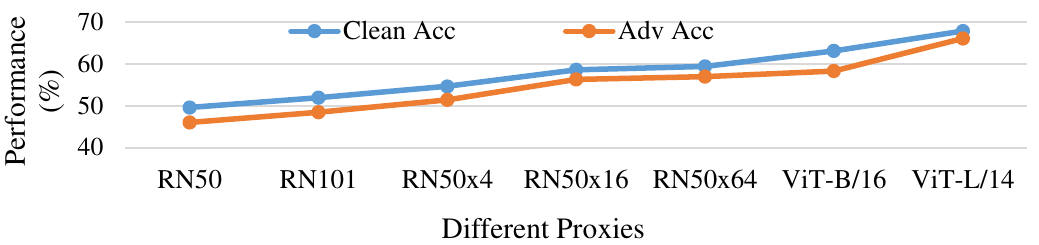}
\caption{The average result on 16 datasets with different CLIPs as a proxy and ViT-B/32-based CLIP as the target model.}
\label{fig_difpro}
\end{figure}

\begin{table}[t]
\centering
\resizebox{0.48\textwidth}{!}
{\begin{tabular}{ccccccccc}
\hline
& \multirow[c]{1}{*}{Model}
&  RN20~\cite{he2016deep} &  RN32 & RN44& VGG16~\cite{simonyan2014very}&  MobileNet~\cite{sandler2018mobilenetv2}  &  ShuffleNet~\cite{zhang2018shufflenet} &  RepVGG~\cite{ding2021repvgg}
\\ \hline
& None   &87.11 &89.80 &91.31 &92.97 &88.10 &84.22 &90.41\\
& RN20 &3.14	&6.54	&8.07	&14.57	&7.50	&13.04	&6.41
\\
& VGG16 &9.97	&8.07	&10.58	&0.99	&10.76	&15.14	&7.22
 \\
\hline
\end{tabular}}
\caption{The proxy adversarial robustness of traditional models. The accuracy of adversarial samples generated by ResNet20 and VGG16 is measured on multiple models.}
\label{tabel6}
\end{table}

\subsection{Principle for Selecting Proxy} Since CLIP has no other variant on some factors (e.g., patch size), targeted ablation studies are infeasible. As an alternative, Fig.~\ref{fig_difpro} shows proxy robustness variations across proxy changes. For CLIPs, model differences are large enough to enable proxy robustness (Adv Acc is close to Clean Acc). We establish an effective proxy selection criterion: better clean generalization, stronger proxy robustness. Notably, a large model gap is not equal to a good proxy: compared with ViT-based CLIP, the ResNet family provides a larger gap but weaker proxy robustness for ViT-B/32-based CLIP, due to their weak generalization.

\subsection{Limitations and Discussion}
Although proxy adversarial robustness is widely present in heterogeneous models of CLIP and beyond CLIP, we do not explain why. To explore this, we verify that there is no similar proxy adversarial robustness for traditional classification models, as shown in Table~\ref{tabel6}. We use 5 models as proxies to verify their performance on adversarial samples generated by ResNet20 (RN20) and VGG16.
Compared with the classification accuracy of natural samples (the first row),  there are poor defense capabilities against adversarial examples generated by heterogeneous networks. This proves that the architectural difference of the model may not be the key to the proxy robustness. Further, we have a simple conjecture: unlike the fixed label space of traditional classification models, VLMs usually update the image and text encoders synchronously during pre-training for contrastive learning. This optimization method allows VLMs to have a wider learning space while ensuring greater differences between heterogeneous models, thereby providing proxy adversarial robustness.

\section{Conclusion}

Regarding the zero-shot adversarial robustness of VLMs, this paper reveals an interesting finding: proxy adversarial robustness. Represented by CLIPs, vanilla CLIP (without adversarial training) can exhibit inherent resilience against adversarial samples generated by its heterogeneous CLIPs. Building on this insight, we propose a heterogeneous proxy transfer strategy via generalization-pivot decoupling (HPT-GPD), which uses the knowledge distillation pipeline between heterogeneous models to improve the robustness of the target model. Experimental results show that on multiple downstream datasets, the proposed HPT-GPD could significantly promote zero-shot adversarial robustness without compromising its natural generalization.

\section{Acknowledgement}

This work was partially supported by National Natural Science Fund of China under Grants 92570110 and 62271090, Chongqing Natural Science Fund under Grant CSTB2024NSCQ-JQX0038, National Key R\&D Program of China under Grant 2021YFB3100800 and National Youth Talent Project.

\bibliographystyle{fcs}
\bibliography{ref}


\begin{biography}{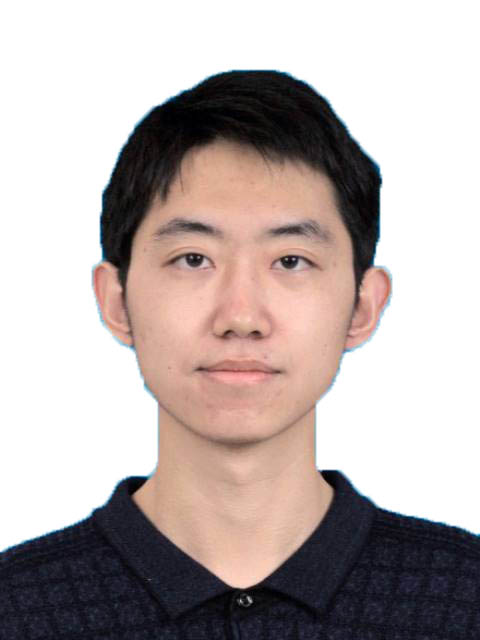}
{Xiaowei Fu}
received the B.S. degree and M.S. degree from Chongqing University, Chongqing, China. He is currently pursuing his Ph.D degree at Chongqing University.

His research interests include computer vision, deep learning, adversarial defense and person ReID.
\end{biography}


\begin{biography}{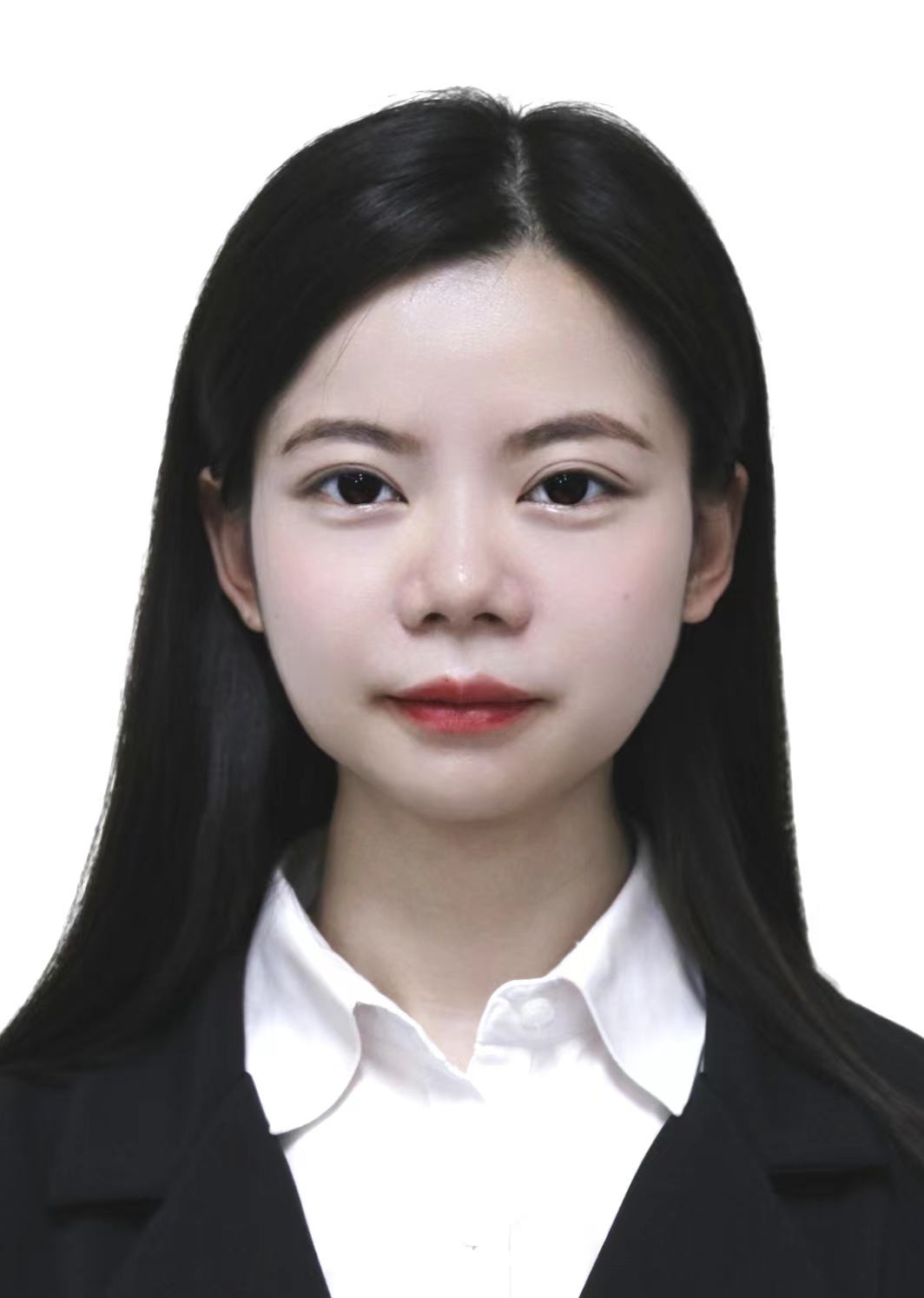}
{Fuxiang Huang} received her Ph.D degree in Information and Communication Engineering from Chongqing University,  Chongqing,  China,  in 2023 and worked as a postdoc at The Hong Kong University of Science and Technology. She is now a research assistant professor at Lingnan Univeristy, Hong Kong. She has published more than 20 technical articles in top journals and conferences,  such as IEEE T-PAMI,  IJCV,  T-IP,  T-NNLS,  T-MM,  T-CSVT,  CVPR and AAAI.
Her current research interests include computer vision,  deep learning,  domain adaptation and multi-modal retrieval.
\end{biography}

\vspace{7mm}

\begin{biography}{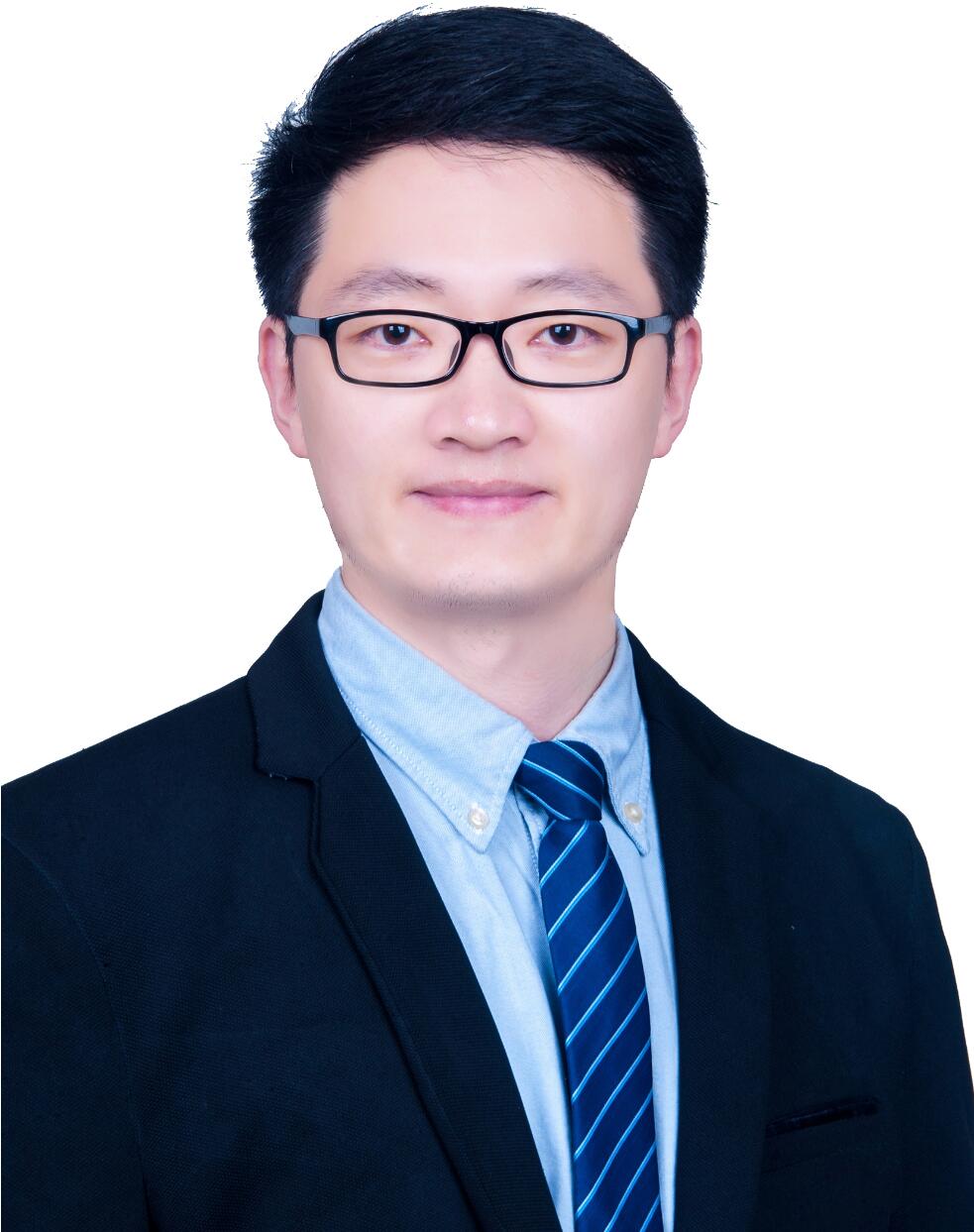}
{Lei Zhang}
received his Ph.D degree in Circuits and Systems from the College of Communication Engineering, Chongqing University, Chongqing, China, in 2013. He worked as a Post-Doctoral Fellow with The Hong Kong Polytechnic University, Hong Kong, from 2013 to 2015. He is currently a Full Professor with Chongqing University and the director of the Chongqing Key Laboratory of Bio-perception \& Multimodal Intelligent Information Processing. He has authored more than 150 scientific papers in top journals and conferences, including IEEE T-PAMI, IJCV, T-IP, T-MM, T-CSVT, T-NNLS, CVPR, ICCV, ECCV, etc. He is on the Editorial Boards of several journals, such as IEEE T-IP, T-IM, Neural Networks, etc.

His current research interests include deep learning, transfer learning, domain adaptation and computer vision.
\end{biography}

\end{document}